\documentclass[10pt,journal,compsoc]{IEEEtran}
\usepackage{times}
\usepackage{epsfig}
\usepackage{graphicx}
\usepackage{amsmath}
\usepackage{amssymb}
\usepackage{algorithm}
\usepackage{algpseudocode}
\usepackage{multirow}
\usepackage{array}
\usepackage{bm}
\usepackage{subfig}
\usepackage{diagbox}
\usepackage{url}

\usepackage{tabularx}

\usepackage[colorlinks, breaklinks=true,bookmarks=false]{hyperref}

\def\etal{\emph{et al.}}

\ifCLASSOPTIONcompsoc
  \usepackage[nocompress]{cite}
\else
  \usepackage{cite}
\fi
\ifCLASSINFOpdf
\else
\fi
\begin{document}
%
\title{Liquid Warping GAN with Attention: A Unified Framework for Human Image Synthesis}

\author{
	Wen~Liu,
	Zhixin Piao, Zhi Tu, Wenhan Luo, Lin Ma, and Shenghua~Gao
	\IEEEcompsocitemizethanks{\IEEEcompsocthanksitem Wen Liu is with School of Information Science and Technology, ShanghaiTech University, and Chinese Academy of  Sciences, Shanghai Institute of  Microsystem and Information Technology, and University of Chinese Academy of Sciences, China.
	\IEEEcompsocthanksitem Zhixin Piao, Zhi Tu, and Shenghua Gao are with School of Information Science and Technology, ShanghaiTech University, Shanghai 201210, China. Shenghua Gao is the corresponding author.\protect\\
	}
	}

\IEEEtitleabstractindextext{%
\begin{abstract}
We tackle human image synthesis, including human motion imitation, appearance transfer, and novel view synthesis, within a unified framework. It means that the model, once being trained, can be used to handle all these tasks. The existing task-specific methods mainly use 2D keypoints (pose) to estimate the human body structure. However, they only express the position information with no abilities to characterize the personalized shape of the person and model the limb rotations. In this paper, we propose to use a 3D body mesh recovery module to disentangle the pose and shape. It can not only model the joint location and rotation but also characterize the personalized body shape. To preserve the source information, such as texture, style, color, and face identity, we propose an Attentional Liquid Warping GAN with Attentional Liquid Warping Block (AttLWB) that propagates the source information in both image and feature spaces to the synthesized reference. Specifically, the source features are extracted by a denoising convolutional auto-encoder for characterizing the source identity well. Furthermore, our proposed method can support a more flexible warping from multiple sources. To further improve the generalization ability of the unseen source images, a one/few-shot adversarial learning is applied. In detail, it firstly trains a model in an extensive training set. Then, it finetunes the model by one/few-shot unseen image(s) in a self-supervised way to generate high-resolution ($512 \times 512$ and $1024 \times 1024$) results. Also, we build a new dataset, namely Impersonator (iPER) dataset, for the evaluation of human motion imitation, appearance transfer, and novel view synthesis. Extensive experiments demonstrate the effectiveness of our methods in terms of preserving face identity, shape consistency, and clothes details. All codes and dataset are available on \url{https://impersonator.org/work/impersonator-plus-plus.html}.
\end{abstract}

\begin{IEEEkeywords}
	Human Image Synthesis, Motion Imitation, Appearance Transfer, Novel View Synthesis,  Generative Adversarial Network, and One/Few-Shot Learning
\end{IEEEkeywords}}

\maketitle
\IEEEdisplaynontitleabstractindextext

%
\IEEEpeerreviewmaketitle

\begin{figure}[t]
	\begin{center}
		\includegraphics[width=\linewidth]{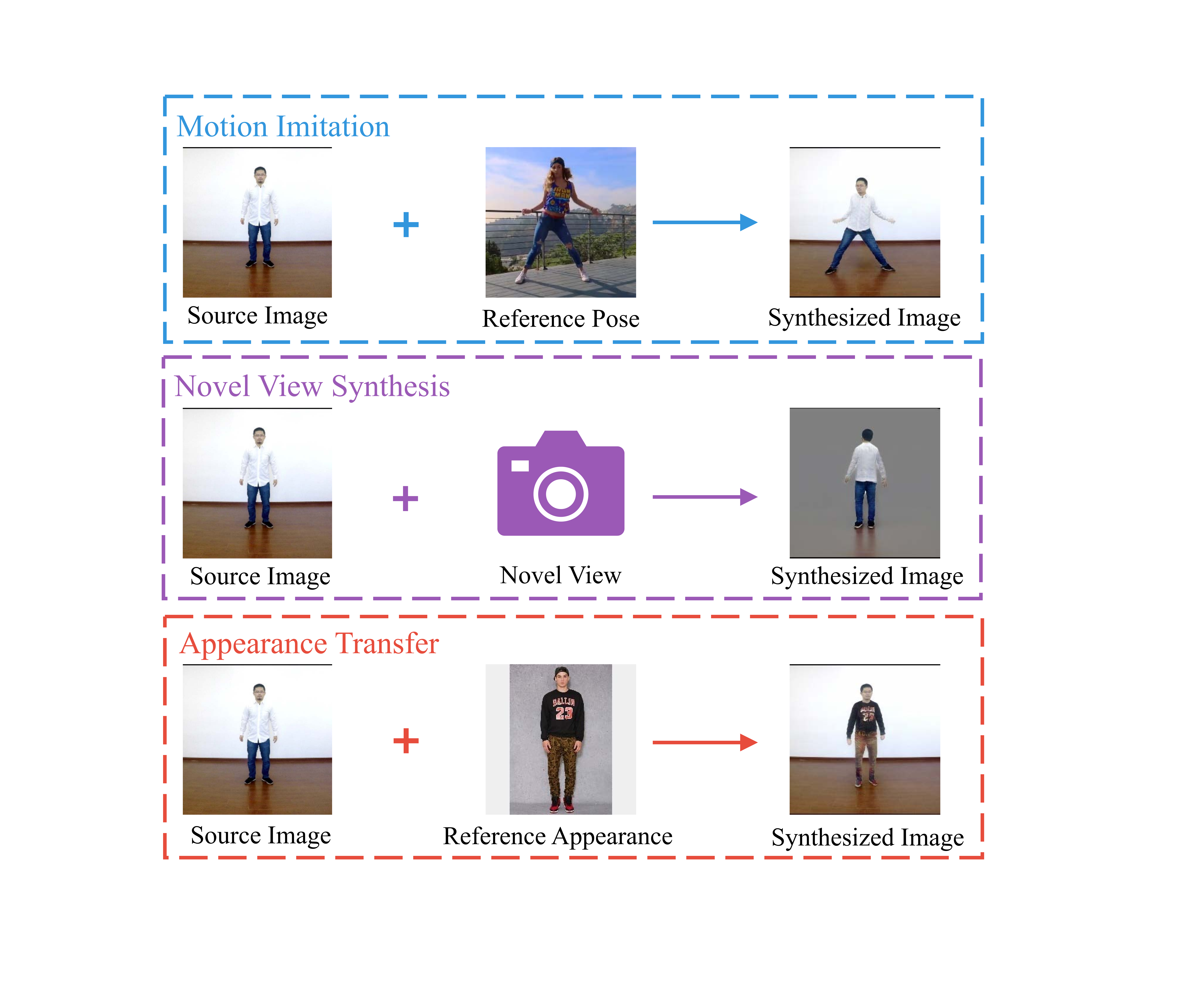}
	\end{center}
	\caption{Illustration of human motion imitation, novel view synthesis and appearance transfer. The $1^{st}$ row is the source image and the $2^{nd}$ row is reference condition, such as image or novel viewpoint of camera. The $3^{rd}$ row is the synthesized results.}
	\label{fig:examples}
\end{figure}

\IEEEraisesectionheading{\section{Introduction}\label{sec:introduction}}
\IEEEPARstart{H}uman image synthesis aims to make believable and photo-realistic images of humans, including motion imitation~\cite{posewarp2018,pG2017nips,DSC2018}, appearance transfer~\cite{swapnet2018,HAT_2018_CVPR} and novel view synthesis~\cite{Zhao0C0JF18,Zhu_2018_CVPR}. It has vast potential applications in character animation, re-enactment, virtual clothes try-on, movie or game making, etc. Given a source human image and a human reference image, i) the goal of motion imitation is to generate an image with the texture from source human and pose from reference human, as depicted in the top row of Fig.~\ref{fig:examples}; ii) human novel view synthesis aims to synthesize new images of the human body, captured from different viewpoints, as illustrated in the middle row of Fig.~\ref{fig:examples}; iii) the goal of appearance transfer is to generate a human image preserving the source face identity while wearing the clothes of the reference, as shown in the bottom row of Fig.~\ref{fig:examples} where each garment (upper-clothes or pants) might come from different people.

\begin{figure}[t]
	\centering
	\includegraphics[width=\linewidth]{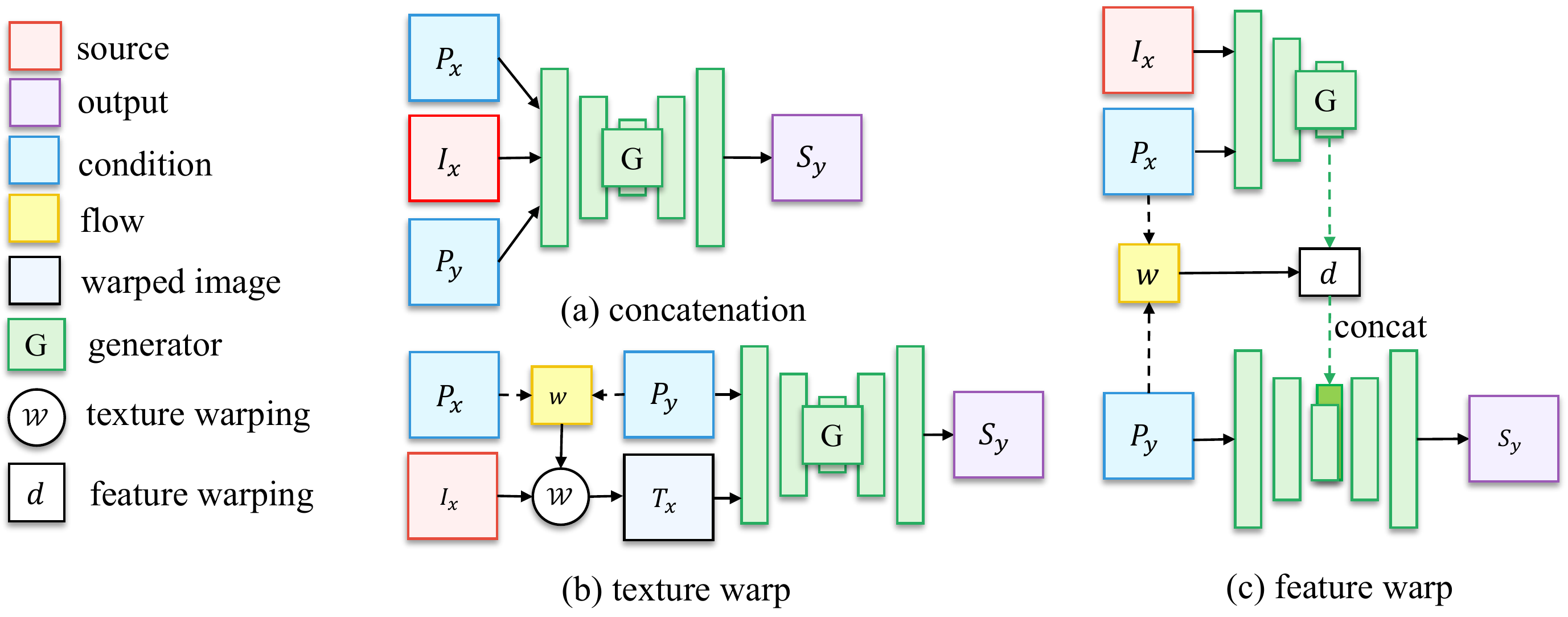}
	\caption{Three existing approaches to propagate the source information into the target condition. (a) early concatenation, concatenates the source image, the source condition, and the target condition into the color channel. (b) and (c) are texture and feature warping, respectively. The source image or its features are propagated into the target condition under a fitted transformation flow.}
	\label{fig:fusion} 
\end{figure}

Taking human motion imitation as an example, existing methods can be roughly categorized into an image-to-image translation-based~\cite{Isola2017ImagetoImageTW,chan2018everybody,Lee_2019_ICLR_MetaPix} pipeline and a warping-based pipeline~\cite{pG2017nips,DSC2018,posewarp2018,softgate18}. The image-to-image translation-based pipeline learns a person-specific mapping function from the human conditions, characterized by a skeleton, dense pose, and parsing result, to the image from a video with paired sequences of conditions and images. Thus, everybody needs to train their model from scratch, and a particular trained model cannot
be applied to others. Besides, it is not accessible to be extended to other tasks, such as appearance transfer. To overcome this shortcoming, researchers have proposed the warping-based methods, which warp the input images into the reference conditions (skeleton, dense pose, or parsing) and generate the desired image. So a trained model in these methods could be applied to other input images with different identities. We summarize the recent warping-based approaches in Fig.~\ref{fig:fusion}. An early work~\cite{pG2017nips}, shown in Fig.~\ref{fig:fusion} (a), feeds the concatenated source image (with its pose condition) with the target pose condition into a network with an adversarial training to generate an image with the desired pose. However, direct concatenation does not consider the spatial layout, and it is ambiguous for the generator to place the pixel from a source image into the right position. Thus, it always results in a blurred image and loses the source identity. Later, inspired by the spatial transformer networks (STN)~\cite{STN2015}, a texture warping method~\cite{posewarp2018}, as shown in Fig.~\ref{fig:fusion} (b), is proposed. It firstly fits a rough affine transformation matrix from the source and the reference key points, then uses an STN to warp the source image into the reference pose, and after that generates the final result based on the warped image. However, texture warping could not preserve the source information as well, in terms of the color, style, or face identity, because the generator might drop out the source information after several downsampling operations, such as stride convolution and pooling. Meanwhile, contemporary work~\cite{softgate18, DSC2018, AlBahar_2019_ICCV} proposes to warp the deep features of the source images into the target poses rather than that in the image space, as shown in Fig~\ref{fig:fusion} (c), named as feature warping.  However, features extracted by an encoder in the feature warping cannot guarantee to characterize the source identity accurately, which consequently produces a blur or low-fidelity image inevitably.

The aforementioned existing methods encounter with challenges in generating realistic-looking images, due to three reasons: 1) diverse clothes in terms of texture, style, color, and high-structure face identity are difficult to be captured and preserved in their network architectures; 2) articulated and deformable human bodies result in a large spatial layout and geometric changes for arbitrary pose manipulations; 3) all these methods cannot handle multiple source inputs, such as in appearance transfer, different parts might come from different source people; 4) the generalization is not good when the inputs are out of the domain of training set because to synthesize photo-realistic images, all these methods apply the adversarial constraints of discriminators, which push the results similar to the distribution of training set.

In this paper, we follow the warping-based pipeline. To preserve the source details of the clothes and face identity, we propose a Liquid Warping Block (LWB) and an advanced version, Attentional Liquid Warping Block (AttLWB), to address the loss of the source information from three aspects: 1) a denoising convolutional auto-encoder is used to extract useful features that preserve the source information, including texture, color, style and face identity; 2) the source features of each local part are blended into a global feature stream by our proposed LWB and AttLWB, to preserve the source details further; 3) it supports multiple-source warping, such as in the appearance transfer that supports to warp the features of a head (local identity) from one source and that of a body from another, and aggregate them into a global feature stream; 4) a one/few-shot learning strategy is utilized to improve the generalization of the network.

In addition, existing approaches mainly rely on a 2D pose~\cite{posewarp2018, pG2017nips, DSC2018}, a dense pose~\cite{DensePoseTransfer} and body a parsing result~\cite{softgate18}. These methods only take care of the layout locations and ignore the personalized shape and limb (joints) rotations, which are even more essential than layout locations in human image synthesis. For example, in an extreme case that a tall man imitates the actions of a short person, if we the 2D skeleton, the dense pose and the body parsing condition will unavoidably change the height and the size of the tall one, as shown at the bottom of Fig.~\ref{fig:comparison}. To overcome these issues, we use a parametric statistical human body model, SMPL~\cite{SMPLify, SMPL:2015, HMR, kolotouros2019spin}, which disentangles a human body into the pose (joint rotations) and the shape. It outputs a 3D mesh (without clothes) rather than the layouts of joints and parts. Further, transformation flows can be easily calculated by matching the correspondences between two 3D triangulated meshes, which is more accurate and results in fewer misalignments than previous fitted affine matrix from keypoints~\cite{posewarp2018, DSC2018}. 

Based on the SMPL model and the Liquid Warping Block (LWB) or the Attentional Liquid Warping Block (AttLWB), our method can be further extended into other tasks, including human appearance transfer and novel view synthesis for free and one model can handle these three tasks. We summarize our contributions as follows: 1) we propose an LWB and an AttLWB to propagate and address the loss of the source information, such as texture, style, color, and face identity, in both the image and the feature space; 2) by taking advantages of both the LWB (AttLWB) and the 3D parametric model, our method is a unified framework for human motion imitation, appearance transfer, and novel view synthesis; 3) since the previous datasets~\cite{liuLQWTcvpr16DeepFashion,fashionvideo_bmvc_2019} have the limitation in the diversity of the poses, and can only be used for motion imitation, we build a dataset for these tasks, especially for human motion imitation in the video, and released all codes and datasets for further research convenience in the community.

This paper is an extension of our previous work~\cite{Liu_2019_ICCV}. We extend the framework in the following aspects: 

i) our previous LWB~\cite{Liu_2019_ICCV} directly adds the warped multiple source features into the global features, and it will enlarge the magnitude of the features in the overlap area, thereby resulting in artifacts. To address this, motivated by the attention architecture~\cite{Vaswa_2017_NeurIPS}, we propose a more advanced Attentional Liquid Warping Block (AttLWB). It firstly learns similarities of the global features among all multiple sources features, and then it fuses the multiple sources features by a linear combination of the learned similarities and the multiple sources in the feature spaces. Finally, to better propagate the source identity (style, color, and texture) into the global stream, we warp the fused source features to the global stream by the Spatially-Adaptive Normalization (SPADE)~\cite{park2019SPADE}, which could further improve the final result;

ii) our previous network could not generalize well when the input images are far away from the training domain, as the interracial motion imitation. The reason might be that to generate images with high fidelity, an adversarial (GAN) loss is essential~\cite{pG2017nips, posewarp2018, DSC2018, Liu_2019_ICCV}, which pushes the generated images in the distribution of the training set. Considering that the input images are diverse in human races, face identities, and clothes styles, and it is infeasible to collect a dataset containing all these individuals. In the testing phase, once an individual is unique in face identity or clothes style, the well-trained network might produce a high-fidelity result similar to the training samples but does not preserve its own source identity in terms of face and clothes. To improve the generalization, inspired by the SinGAN~\cite{Shaham_2019_ICCV} and the Few-Shot Adversarial Learning~\cite{Zakharov_2019_ICCV}, we apply a one/few-shot adversarial learning to push the network to focus on the individual input with several steps of adaptation, namely personalization.

iii): our previous method successfully achieves decent results on $256 \times 256$ resolution, and in this version, based on the AttLWB and personalization, we could further achieve the high-fidelity results with a higher $512 \times 512$ and $1024 \times 1024$ resolution.

We organize the rest of this paper as follows: In Section 2, we summarize the related work of the Human Image Synthesis, including the motion imitation,  the appearance transfer, and the novel view synthesis.
In Section 3, we firstly introduce the essential modules of our proposed Attentional Liquid Warping GAN. The following are the training strategies, the loss functions, the one/few-shot personalization, and the inference details. In Section 4, extensive experiments on different datasets and tasks validate the effectiveness of our work. In Section 5, ablation studies and analysis are conducted to evaluate the impacts of different components. We conclude our work in Section 6.

\begin{figure*}[t] 
	\centering
	\includegraphics[width=1.0\linewidth]{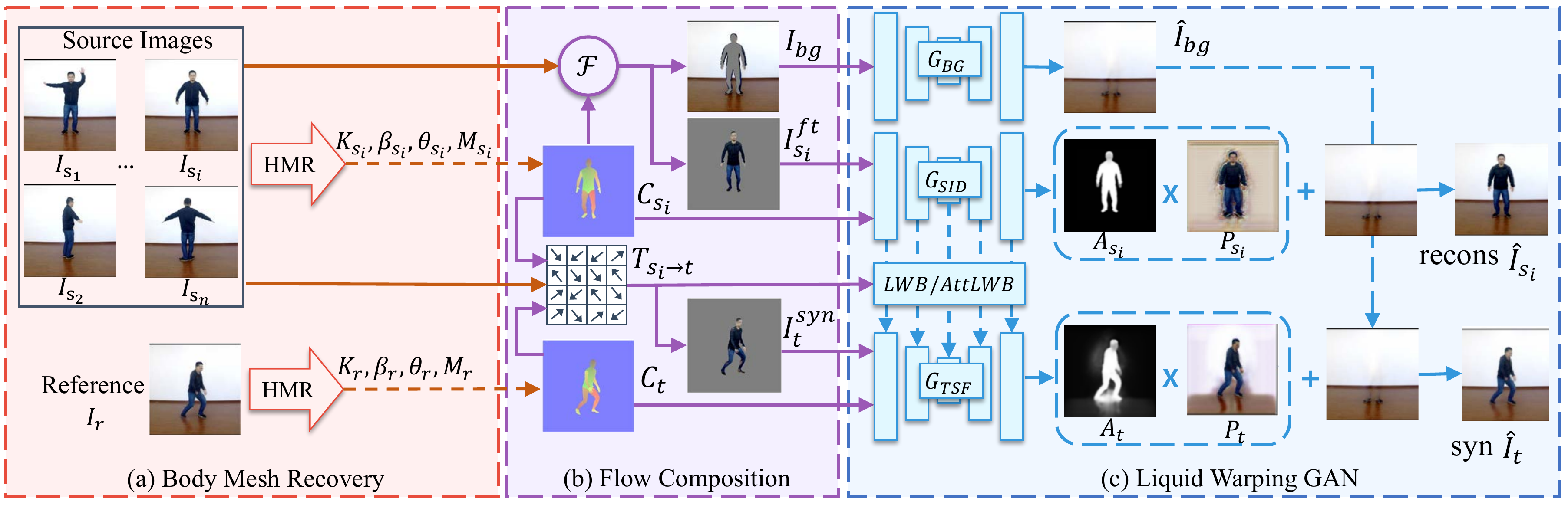}   
	\caption{The training pipeline of our method. We randomly sample a pair of images from a video, denoting the source and the reference image as $I_{s_i}$ and $I_r$. \textbf{(a)} A body mesh recovery module will estimate the 3D mesh of each image and render their correspondence map, $C_s$ and $C_t$; \textbf{(b)} The flow composition module will first calculate the transformation flow $T$ based on two correspondence maps and their projected vertices in the image space. Then it will separate the source image $I_{s_i}$ into a foreground image $I^{ft}_{s_i}$ and a masked background $I_{bg}$. Finally it warps the source image based on the transformation flow $T$ and produces a warped image $I_{syn}$; \textbf{(c)} In the last GAN module, the generator consists of three streams, which separately generates the background image $\hat{I}_{bg}$ by $G_{BG}$, reconstructs the source image $\hat{I}_s$ by $G_{SID}$ and synthesizes the target image $\hat{I}_t$ under the reference condition by $G_{TSF}$. To preserve the details of the source image, we propose a novel LWB and AttLWB (shown in Fig.~\ref{fig:lwb}) which propagates the source features of $G_{SID}$ into $G_{TSF}$ at several layers and preserve the source information, in terms of texture, style and color.}
	\label{fig:pipeline} 
\end{figure*}

\section{Related work}\label{related_work}

\subsection{Human Motion Imitation}
We summarize the recent image-to-image translation-based and the warping-based methods as follows.

\textbf{Image-to-Image translation-based methods.} Esser~\etal\cite{vunet2018} use a Variational U-Net to learn a mapping function from a 2D skeleton to an image. Chan~\etal\cite{chan2018everybody} learn a mapping function from a 2D skeleton to an image by a pix2pixHD~\cite{Wang_2018_CVPR} with a specialized Face GAN and temporally coherent GAN. Wang~\etal\cite{wangVID2VID} propose a vid2vid framework and learn a mapping function from 2D dense pose to image. Meanwhile, Shysheya~\etal\cite{Shysheya_2019_CVPR} firstly build a full texture UV image of a person by multi-view cameras, then learn a mapping function from a 3D skeleton to part coordinates of the UV map and finally render a result based on the coordinates and the UV image. Contemporarily, Liu~\etal\cite{Liu2018Neural} firstly use a monocular video to reconstruct a full 3D character model of a person with a static pose, then render the texture of each body parts and finally learn a mapping from synthetic to real images. However, all these methods train a mapping from keypoints or parts to each person's image and everybody needs to train their own model. This might limit its wide application. 

\textbf{Warping-based methods.} Recent work is mainly based on the conditioned generative adversarial networks (CGAN)~\cite{pG2017nips, posewarp2018, ma2018disentangled, DensePoseTransfer, Si_2018_CVPR}. Their key technical idea is to combine the source image along with the source pose (2D skeleton) as inputs and generate a realistic image by GANs using a reference pose. The differences among those approaches are merely in network architectures, warping strategies, and adversarial losses. In~\cite{pG2017nips}, Ma~\etal\cite{pG2017nips} directly concatenate the source image and the reference pose, and then design a U-Net~\cite{unet2015} generator with a coarse-to-fine strategy to generate $256\times256$ images. Neverova~\etal~\cite{DensePoseTransfer} replace the sparse 2D key points with the dense correspondences between the image and surface of the human body by the DensePose~\cite{DensePose}. Si~\etal~\cite{Si_2018_CVPR} propose a multistage adversarial loss and separately generate the foreground (or different body parts) and background. Balakrishnan~\etal\cite{posewarp2018} firstly fit an affine transformation matrix based on the source and the target 2D key points and then use a texture warping strategy to generate the foreground and the background separately. These work~\cite{DSC2018, softgate18, Zhu_2019_CVPR, AlBahar_2019_ICCV}, focus on the way of warping the source features into the target conditions, like skeleton or parsing. Besides, Li~\etal\cite{Li_2019_CVPR} propose to learn a transformation flow from 2D key points and warp the deep features based on the learned transformations.

\subsection{Human Appearance Transfer} Human appearance modeling or transfer is a vast topic, especially in the field of virtual try-on applications, from computer graphics pipelines~\cite{ponsmollSIGGRAPH17clothcap} to learning based pipelines~\cite{swapnet2018, HAT_2018_CVPR}. Graphics based methods first estimate the detailed 3D human mesh with clothes via garments and 3D scanners~\cite{3dScanZhang17} or multiple camera arrays~\cite{mvLeroyFB17}, and then human appearance with clothes is capable of being conducted from one person to another based on the detailed 3D mesh. Although these methods can produce high-fidelity results,
their cost, size, and controlled environment are unfriendly and inconvenient to customers. Recently, in the light of deep generative models, SwapNet~\cite{swapnet2018} firstly learns a pose-guided clothing segmentation synthetic network, and then the clothing parsing results with texture features from the source image are fed into an encoder-decoder network to generate the image with the desired garment. In~\cite{HAT_2018_CVPR}, the authors leverage a geometric 3D shape model combined with learning methods, swap the color of visible vertices of the triangulated mesh, and train a model to infer that of invisible vertices. Instead of estimating the 3D clothes by other sensors, in the MGN~\cite{Bhatnagar_2019_ICCV}, the authors, train a network with 3D scans data and predict the body shape and clothing directly from 8 frames or a video. They apply the garment transfer based on the estimated 3D body mesh with clothes.

\subsection{Human Novel View Synthesis} Novel view synthesis aims to synthesize new images of the same object or human body from arbitrary viewpoints. The core step of existing methods is to fit a correspondence map from the observable views to new views with convolutional neural networks. In~\cite{ZhouTSME16}, the authors use CNNs to predict appearance flow and synthesize new images of the same object by copying the pixel from a source image based on the appearance flow and they have achieved decent results of rigid objects like vehicles. The following work~\cite{Park_2017_CVPR} proposes to infer the invisible textures based on appearance flow and adversarial generative network (GAN)~\cite{gan2014}, while Zhu~\etal~\cite{Zhu_2018_CVPR} argue that appearance flow-based method performs poorly on articulated and deformable objects, such as human bodies. They propose an \emph{appearance-shape-flow} strategy to synthesize different views of human bodies -- besides, Zhao~\etal~\cite{Zhao0C0JF18} design a GAN based method to synthesize high-resolution views in a coarse-to-fine way. Recently, in PiFu~\cite{Saito_2019_ICCV}, the authors learn an implicit function with multi-layer perceptrons (MLPs) to digitize the human body and infer the 3D surfaces and texture from a single or multiple frames. The fully digitalized human body could synthesize a different view.

\subsection{One/Few-shot Learning in Image Synthesize}
Ding~\etal~\cite{Ding_2018_OneShotFace} propose a generative adversarial one-shot face recognizer to synthesize new face images. Shaham~\etal  ~\cite{Shaham_2019_ICCV} introduce a SinGAN, an unconditional generative model from a single image. Zakharov~\etal~\cite{Zakharov_2019_ICCV} apply the few-shot adversarial learning to generate the realistic talking head. In light of the success of the Meta-Learning in classification, reinforcement learning and network architecture search~\cite{Finn_2017_MAML_ICML, Nichol_2018_Reptile,lian2020iclr}, Lee~\etal~\cite{Lee_2019_ICLR_MetaPix} propose a MetaPix for the few-shot motion imitation. Wang~\etal~\cite{Wang_2019_NeurIPS} extend the previous vid2vid~\cite{wangVID2VID} framework within a few-shot setting and make it capable of synthesizing videos of unseen subjects by leveraging few example images.

\section{Our Approach}
In this section, we first introduce the whole models of our framework. It contains three modules, a body mesh recovery, a flow composition, and a GAN module with the Liquid Warping Block (LWB) or the Attentional Liquid Warping Block (AttLWB). Then, the following are the training details and loss functions. Further, to improve the generalization, we introduce a one/few-shot learning strategy. We illustrate the details of how to apply our model to three tasks in the inference section (Sect.~\ref{bookmark:inference}).

Once the model has been trained on one task, it can deal with other tasks as well. Here, we use motion imitation as an example, as shown in Fig. 3. Our framework supports multiple sources of inputs, denoting the source images as $\{I_{s_1}, I_{s_2}, ..., I_{s_n}\}$, and the reference image as $I_r$. Here, $s_n$ is the number of source images. First, the body mesh recovery module will estimate the 3D mesh of $I_{s_i}$ and $I_r$ and render their correspondence maps, $C_{s_i}$, and $C_t$. Next, the flow composition module will calculate the transformation flow $T_{s_i\to t}$ of each source image to the reference, based on two correspondence maps and their projected mesh in image space. Each source image $I_{s_i}$ is thereby decomposed as the foreground image $I^{ft}_{s_i}$ and the masked background $I^{bg}_{s_i}$.  Since all source images share the same background, we randomly choose one of the masked backgrounds, denoted as $I_{bg}$. Simultaneously, 
each source image contributes its visible textures to warp a synthetic image $I^{syn}_t$, based on the transformation flow $T_{s_i\to t}$. The last (Attentional) Liquid Warping GAN module consists of three streams. It separately generates the background image by $G_{BG}$, reconstructs the source image $\hat{I}_{s_i}$ by $G_{SID}$ and synthesizes the final result $\hat{I_t}$ under the reference condition by $G_{TSF}$. To preserve the details of source image, we propose the novel Liquid Warping Block (LWB) and Attentional Liquid Warping Block (AttLWB) which propagate the source features of $G_{SID}$ into $G_{TSF}$ at multiple layers.

\subsection{Body Mesh Recovery Module}
As shown in Fig.~\ref{fig:pipeline} (a), given the source image $I_{s_i}$ and the reference image $I_r$, the role of this stage is to predict the kinematic pose (rotation of limbs) and shape parameters, as well as the 3D mesh of each image. In this paper, we use the HMR~\cite{HMR, kolotouros2019spin} as the 3D pose and shape estimator due to its good trade-off between accuracy and efficiency. In HMR, an image is firstly encoded into a feature with $\mathbb{R}^{2048}$ by a ResNet-50~\cite{resnetv2He16} and then followed by an iterative 3D regression network that predicts the pose $\theta \in\mathbb{R}^{72}$ and the shape $\beta \in\mathbb{R}^{10}$ of SMPL~\cite{SMPL:2015}, as well as the weak-perspective camera $K\in\mathbb{R}^3$. SMPL is a 3D body model that can be defined as a differentiable function $M(\theta, \beta) \in \mathbb{R}^{N_v \times 3}$, and it parameterizes a triangulated mesh by $N_v = 6,890$ vertices and $N_f = 13,776$ faces with the parameters of a pose $\theta \in\mathbb{R}^{72}$ and a shape $\beta \in\mathbb{R}^{10}$. Here, the shape parameters $\beta$ are the coefficients of a low-dimensional shape space learned from thousands of registered scans, and the pose parameters $\theta$ are the joint rotations that articulate the bones via forwarding kinematics. With such process, we will obtain the body reconstructive estimations of each source image, $\{K_{s_i}, \theta_{s_i}, \beta_{s_i}, M_{s_i}\}$ and those of reference image, $\{K_r, \theta_r, \beta_r, M_r\}$, respectively.

\subsection{Flow Composition Module}
\label{sec:stage2}
Based on previous estimations, we first render a correspondence map and a weight index map for each source mesh $M_{s_i}$ and the reference mesh $M_r$ under the camera view of $K_{s_i}$ and $K_r$. Here, we denote the source weight index map, the source and the target correspondence maps as $W_{s_i}$, $C_{s_i}$ and $C_t$, respectively. In this paper, we use a fully differentiable renderer, Neural Mesh Renderer (NMR)~\cite{cvprKatoUH18}. 
We thereby project vertices of the source $V_{s_i}$ into a 2D image space by a weak-perspective camera, $v_{s_i}=\pi(V_{s_i}, K_{s_i})$. Here, $\pi$ is the weak-perspective projective function.  Then, we calculate the barycentric coordinates of each mesh face and obtain $f_{s_i} \in \mathbb{R}^{N_f \times 2} $. Next, we calculate the transformation flow $T_{s_i \to t}\in\mathbb{R}^{H\times W\times 2}$ by matching the correspondences between the source correspondence map with its mesh face coordinates $f_{s_i}$. Here $H\times W$ is the size of the image. By the same means, we obtain the transformation flow $T_{r\to t}$ of the reference correspondence map. We describe the procedure to obtain the transformation flow in Algorithm~\ref{alg:flow}.
Consequently, a foreground image $I^{ft}_{s_i} $ and a masked background image $I^{bg}_{s_i}$ are derived from masking the source image $I_{s_i}$ based on $C_{s_i}$. We randomly pick one of the masked backgrounds, denoted as $I_{bg}$, because all source images share the same background. Finally, we warp the visible textures of each source image $I_{s_i}$ to the desired condition by the transformation flow $T_{s_i \to t}$ and thereby obtain a synthetic image $I^{syn}_t$, as depicted in Fig.~\ref{fig:pipeline}.

\begin{algorithm}[htb] 
	\caption{The procedure of obtaining transformation $T_{s_i \to t}$.} 
	\begin{algorithmic}[1] 
		\Require $W_{s_i}$, $V_{s_i}$, $F_{s_i}$, $K_{s_i}$, $C_{s_i}$, $C_{t}$.
		
		\begin{itemize}
			\item $K_{s_i}\in \mathbb{R}^{3 \times 1}$: source weak-perspective camera;
			\item $V_{s_i} \in \mathbb{R}^{N_v \times 3}$: $N_v$ is the number of vertices;
			\item $F_{s_i} \in \mathbb{R}^{N_f \times 3}$: $N_f$ is the number of faces;
			\item $W_{s_i} \in \mathbb{R}^{H \times W \times 3}$: the weight index map of source mesh, the value of each pixel indicates the barycentric weights of the triangulated faces in image space;
			\item $C_{s_i} (C_t) \in \mathbb{R}^{H \times W \times 1}$: the correspondence map of source and target mesh, and the value in each pixel indicates the face index of the mesh.
		\end{itemize}
		
		\Ensure $T_{s_i \to t} \in \mathbb{R}^{H \times W \times 2}$, the output transformation flow; 
		\State $v_{s_i} = \pi(V_{s_i}, K_{s_i})$ \# projecting vertices of source $V_{s_i}$ into the 2D image space by the weak-perspective camera;
		\label{code:getvs}
		\State $tri_{s_i} = v_{s_i}[F_{s_i}]$ $\in \mathbb{R}^{N_f \times 3 \times 2}$  \# the triangulated faces with vertices in 2D image space;
		\State $Vis_{s_i} \in \mathbb{R}^{N_f \times 1}$ \# the face visibility;
		\For{$f=1$ to $N_f$} 
		\State $Vis_{s_i}(f) = 1$ if $f$ appears in $C_{s_i}$ else 0; 
		\EndFor 
		\State initializing $T_{s_i \to t} \in \mathbb{R}^{H \times W \times 2}$;
		\For{$i=1$ to $H$}
		\For{$j=1$ to $W$}
		\State $f = C_t(i, j)$ \# the face index in current pixel;
		\State $T_{s_i \to t}(i, j) = W_{s_i}(i,j) \times tri_{s_i}(f)$, if $Vis_{s_i}(f)$ is 1;
		\EndFor
		\EndFor 
		\\ 
		\Return $T_{s_i \to t}$.
	\end{algorithmic} 
	\label{alg:flow}
\end{algorithm}

\subsection{Attentional Liquid Warping GAN}
This stage synthesizes high-fidelity human images under the desired condition. More specifically, it 1) synthesizes the background image; 2) predicts the color of invisible parts based on the visible parts; 3) generates pixels of clothes, hairs, and others out of the reconstruction of SMPL.

\textbf{Generator.} Our generator works in a three-stream manner. One stream, named $G_{BG}$, works on the concatenation of the masked background image $I_{bg}$ and the mask obtained by the binarization of $C_{s_i}$ in the color channel to generate the realistic background image $\hat{I}_{bg}$, as shown in the top stream of Fig.~\ref{fig:pipeline} (c). The other two streams are the source identity stream, namely $G_{SID}$ and the transfer stream, namely $G_{TSF}$. $G_{SID}$ is a denoising convolutional auto-encoder that aims to guide the encoder to extract the features that are capable of preserving the source information. Together with the $\hat{I}_{bg}$, it takes the masked source foreground $I^{ft}_{s_i}$ and the correspondence map $C_{s_i}$ as its inputs and reconstructs source foreground image $\hat{I}_s$. $G_{TSF}$ stream synthesizes the final result, which receives the warped foreground by a bilinear sampler and the correspondence map $C_t$ as its inputs. To preserve the source information, such as texture, style, and color, we propose a novel Liquid Warping Block (LWB), as well as its advanced version, Attentional Liquid Warping Block (AttLWB), that links the source with the target streams. They blend the source features from $G_{SID}$ and fuses them into the transfer stream $G_{TSF}$, as shown at the bottom of Fig.~\ref{fig:pipeline} (c).

$G_{BG}$ and $G_{SID}$ have similar architectures with separate parameters and follow the structure of CycleGAN~\cite{CycleGAN2017} with 6 residual blocks~\cite{resnetHe16}.  The details of kernel sizes and number of filters are illustrated in Fig.~\ref{fig:network}. $G_{TSF}$ is a combination of a ResNet and a U-Net~\cite{unet2015}, named ResUnet. For $G_{BG}$, we directly regress the final background image, $\hat{I}_{bg}$, while for $G_{SID}$ and $G_{TSF}$, we concretely generate an attention map $A$ and a color map $P$, as shown in Fig.~\ref{fig:network}. The final image can be obtained as follows:
\begin{equation}
\begin{aligned}
\hat{I}_{s_i} &= P_s \odot A_{s_i} + \hat{I}_{bg} \odot (1 - A_{s_i}) \\ 
\hat{I}_t &= P_t \odot A_t + \hat{I}_{bg} \odot (1 - A_t).
\end{aligned}
\label{equ:att_color}
\end{equation}
Here, $\odot$ represents an element-wise multiplication.
The total trainable parameters in the generator are $\theta_G=\{\theta_{BG}, \theta_{SID}, \theta_{TSF}, \theta_{AttLWB}\}$, with respect to $G_{BG}$, $G_{SID}$, $G_{TSF}$ and AttLWB.

\textbf{Discriminator.} To push the discriminators to focus on different aspects of the generated images, such as the clothes on the human body and the face identity, we utilize a global-local content-orientation architecture. It consists of three sub-discriminators. The first one is a global discriminator, $D_{Global}$, which regularizes the entire generated $\hat{I}_t$ to be more realistic-looking. The rest two are a body discriminator $D_{Body}$ and a face discriminator $D_{Head}$, and they push the cropped body area and the head (face) parts of the generated $\hat{I}_t$ to be realistic-looking.  All of them are conditional discriminators, and they take the generated images and the correspondence map $C_t$ as their inputs.  We illustrate the details of our discriminators in Fig.~\ref{fig:network}. The total trainable parameters in the discriminators are $\theta^D = \{\phi_{Global}, \phi_{Body}, \phi_{Head}\}$.

\begin{figure*}[t]
	\begin{center}
		\includegraphics[width=\linewidth]{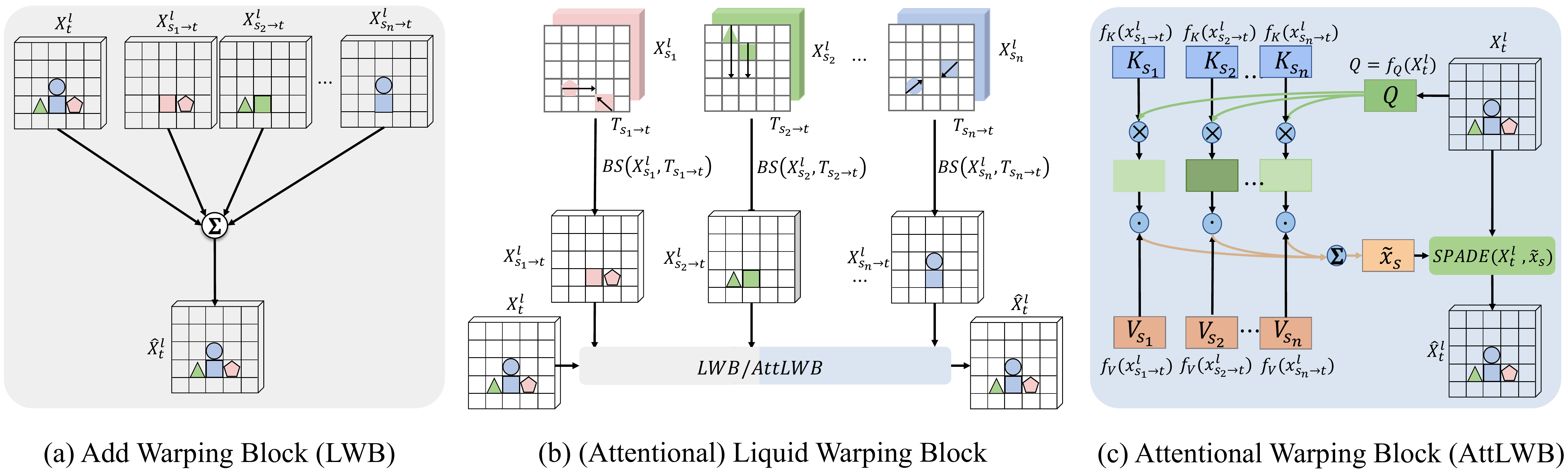}
	\end{center}
	\caption{Illustration of our LWB and AttLWB. They have the same structure illustrated in \textbf{(b)} but with separate AddWB (illustrated in \textbf{(a)}) or AttWB (illustrated in \textbf{(b)}). \textbf{(a)} is the structure of AddWB. Through AddWB, $\widehat{X}_t^{l}$ is obtained by  aggregation of warped source features and features from $G_{TSF}$. \textbf{(b)} is the shared structure of (Attentional) Liquid Warping Block. $\{X^{l}_{s_1}, X^{l}_{s_2}, ..., X^{l}_{s_n}\}$ are the feature maps of different sources extracted by $G_{SID}$ at the $l^{th}$ layer. $\{T_{s_1\to t}, T_{s_2\to t},...,T_{s_n\to t}\}$ are the transformation flows from different sources to the target. $X^{l}_t$ is the feature map of $G_{TSF}$ at the $l^{th}$ layer. \textbf{(c)} is the architecture of AttWB. Through AttWB, final output features $\widehat{X}_t^{l}$ is obtained with SPADE by denormalizing feature map from $G_{TSF}$ with weighted combination of warped source features by a bilinear sampler (BS) with respect to corresponding flow $T_{s_i\to t}$. }
	\label{fig:lwb}
\end{figure*}

\textbf{Attentional Liquid Warping Block.}  One advantage of our proposed Liquid Warping Block (LWB) and Attentional Liquid Warping Block (AttLWB)  is that it addresses the issue of multiple sources. For instance, in human motion imitation, the source images are multi-view inputs, and in the appearance transfer, different parts of garments come from different people. The different parts of features are aggregated into $G_{TSF}$ by their transformation flow independently. As shown in Fig.~\ref{fig:lwb}, we denote $X^{l}_{s_1}$ and $X^{l}_{s_2}$ as the feature maps extracted by $G_{SID}$ of different sources at the $l^{th}$ layer and $X^{l}_t$ is the feature map of $G_{TSF}$ at the $l^{th}$ layer. Each part of the source feature is warped by their transformation flow and aggregated into the features of $G_{TSF}$. We use a bilinear sampler (BS) to warp the source features $X^{l}_{s_1}$ and $X^{l}_{s_2}$ with respect to corresponding transformation flows, $T_{s_1\to t}$ and $T_{s_2 \to t}$.  The way to aggregate the warped source features into the global stream is the main difference between LWB and AttLWB.

LWB, as illustrated in Fig.~\ref{fig:lwb} (a), directly uses an element-wise addition among all features and the fuses the global features as:
\begin{equation}
\begin{aligned}
X^{l}_{s_i \to t} &= BS(X^l_{s_i}, T_i)  \\
\widehat{X}_t^{l} &= \sum_{i=1}^{s_n}X^{l}_{s_i \to t} + X_t^{l}.
\end{aligned}
\label{equ:addlwb}
\end{equation}

However, LWB will enlarge the magnitude of the features in the overlap area, and thereby result in artifacts. To address this, motivated by the attention architecture~\cite{Vaswa_2017_NeurIPS}, we propose a more advanced Attentional Liquid Warping Block (AttLWB), as shown in Fig.~\ref{fig:lwb} (c). It firstly learns similarities of the global features among all multiple source features, and then it fuses the multiple source features by the linear combination of the learned similarities and the multiple sources in feature space. Finally, to better propagate the source identity (style, color, and texture) into the global stream, we use the SPADE~\cite{park2019SPADE} to denormalize the feature map of $G_{TSF}$ with the fused source features to obtain the global stream, which could further improve the final result.  We describe the entire procedures of AttLWB in Algorithm~\ref{alg:attlwb}.
\begin{algorithm}[htb] 
	\caption{The procedure of our AttLWB.} 
	\begin{algorithmic}[1] 
		\Require $\{T_{s_1 \to t}, ..., T_{s_n\to t}\}$, $\{X^{l}_{s_1}, ..., X^{l}_{s_n}\}$, and $X^{l}_t$. 
		
		\begin{itemize}
			\item $\{T_{s_1\to t}, ..., T_{s_n\to t}\}$: the transformation flows from different sources to the target; 
			\item $\{X^{l}_{s_1}, ..., X^{l}_{s_n}\}$: the feature maps extracted by $G_{SID}$ of different sources at the $l^{th}$ layer; 
			\item $X^{l}_t$: the feature map of $G_{TSF}$ at the $l^{th}$ layer;
		\end{itemize}
		
		\Ensure $\widehat{X}_t^{l}$, the output features; 
		\State $X^{l}_{s_i \to t} = BS(X^l_{s_i}, T_{s_i\to t})$ \# warping each source feature; 
		\label{code:fram:warp_feature} 
		\State $Q = f_Q(X^{l}_t)$ \# query embeddings; 
		\label{code:fram:query} 
		\State $K = [f_K(X^{l}_{s_1 \to t}), ..., f_K(X^{l}_{s_n \to t})]$ \# key embeddings;
		\label{code:fram:key} 
		\State $V = [f_V(X^{l}_{s_1 \to t}), ..., f_V(X^{l}_{s_n \to t})]$ \# value embeddings;
		\label{code:fram:value}
		\State $\tilde{x}_s=Attention(Q, K, V) = Softmax(\frac{QK^T}{\sqrt{d_k}})V$ \# fused source features, $d_k$ is the number of channels of $K$;
		\label{code:fram:attention}
		\State $\widehat{X}_t^{l} = SPADE(X^{l}_t, \tilde{x}_s)$ \# conditioned on $\tilde{x}_s$;
		\label{code:fram:adain}  \\ 
		\Return $\widehat{X}_t^{l}$; 
	\end{algorithmic} 
	\label{alg:attlwb}
\end{algorithm}

\begin{figure*}[h]
	\centering
	\includegraphics[width=\linewidth]{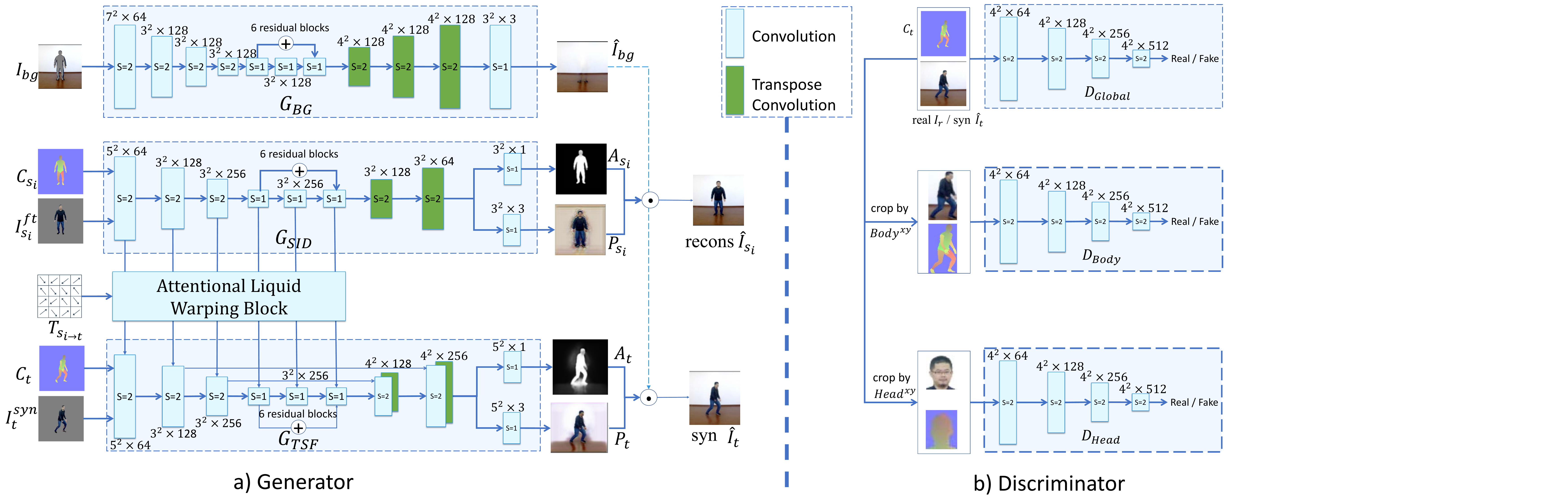}
	\caption{The details of network architectures of our Attentional Liquid Warping GAN, including the generator and the discriminator. Here $s$ represents the stride size in convolution and transposed convolution.}
	\label{fig:network} 
\end{figure*}

\subsection{Training Details and Loss Functions}
In this part, we will introduce the loss functions and how to train the whole system. For the body recovery module, we follow the network architecture and loss functions of HMR~\cite{HMR,kolotouros2019spin}. Here, we use a pre-trained (off-the-shelf) SMPL estimator.

Note that our proposed Attentional Liquid Warping GAN is a unified framework for motion imitation, appearance transfer, and novel view synthesis. Therefore once we have trained the model on one task, it is capable of being applied to other tasks.  These three tasks share the same training pipeline in our method, except for the way to sample the source the reference images. In motion imitation, we randomly sample $s_n + 1$ images from each video with difference poses and set the first $s_n$ ones as the source images $\{I_{s_1}, ..., I_{s_n} \}$ and the other one as the reference $I_r$. In appearance transfer, we need to sample $s_n + 1$ images with the same person identity wearing different clothes, while in novel view synthesis, we need to sample $s_n + 1$ images of the same person under the different camera of views. In our experiments, we train a model for motion imitation and then apply it to appearance transfer and novel view synthesis.

The whole loss function of the generator contains four terms, which are perceptual loss~\cite{eccvJohnsonAF16}, face identity loss, attention regularization loss, and adversarial loss.

\textbf{Perceptual Loss.} It regularizes the reconstructed source image $\hat{I}_{s_i}$ to the ground truth $I_{s_i}$ and pushes the generated target image $\hat{I}_t$ and the reference image $I_r$ to be closer in a  VGG~\cite{Simonyan15} feature subspace. Its formulation is given as follows:   
\begin{equation}
\begin{aligned}
L_p=\frac{1}{s_n}\sum_{i=1}^{s_n}\|\hat{I}_{s_i} - I_{s_i}\|_1 + \|f(\hat{I}_t) - f(I_r)\|_1.
\end{aligned}
\label{equ:percep}
\end{equation}
Here, $f$ is a pre-trained VGG-19~\cite{Simonyan15} on ImageNet~\cite{ILSVRC15}.

\textbf{Face Identity Loss.} It regularizes the cropped face from the synthesized target image $\hat{I}_t$ to be similar to that from the image of ground truth $I_r$, which pushes the generator to preserve the face identity. It is shown as follows: 
\begin{equation}
\begin{aligned}
L_f=\|g(\hat{I}_t) - g(I_r)\|_1.
\end{aligned}
\label{equ:face}
\end{equation}
Here, $g$ is a pre-trained SphereFaceNet~\cite{cvprLiuWYLRS17}.

\textbf{Adversarial Loss.} It pushes the distribution of synthesized images to the distribution of real images. We use a $LSGAN_{-110}$~\cite{lsgan_gp} loss in a way like PatchGAN over all discriminators, $D_{Global}$, $D_{Body}$ and $D_{Head}$. They push the entire generated images, cropped body area, and head (face) parts to be realistic-looking. 
We denote the bounding box of head and body as $head^{xy}$ and $body^{xy}$ in the ground-truth $I_r$, respectively, and we calculate them by the projected vertices in the image space. $\hat{I}^{b}_t$, $I^{b}_r$ and $C^{b}_t$ are the cropped bodies from the generated image, the reference image and the correspondence map, based on bounding box of body, $body^{xy}$. $\hat{I^{h}_t}$, $I^{h}_r$ and $C^{h}_t$ are the corresponding cropped heads with respect to the bounding box of head, $head^{xy}$.
We arrive at the total adversarial loss as follows: 
\begin{equation}
\begin{aligned}
L^G_{adv}&=\sum D_{Global}(\hat{I}_t, C_t)^2 + \sum D_{Body}(\hat{I}^b_t,, C^b_t)^2 \\
&+ \sum D_{Head}(\hat{I}^h_t, C^h_t)^2 
\end{aligned}
\label{equ:adv}
\end{equation}

\textbf{Attention Regularization Loss.} It regularizes the attention map $A_t$ and $A_{s_i}$ to be smooth and prevents them from saturating. Considering that there is no ground truth of attention map $A$ or color map $P$, they are learned from the resulting gradients of above losses. However, the attention masks can easily saturate to 1 which prevents the generator from working. To alleviate this situation, we regularize the mask to be closer to the silhouettes $S$ rendered from a 3D body mesh. Since the silhouettes is a rough map and it contains the body mask without clothes and hair, we addtionaly introduce a Total Variation Regularization~\cite{ganimation} over $A$ to compensate the shortcomings of silhouettes. It is shown as:
\begin{equation}
\small
\begin{aligned}
L_a&= \|A_s-S_s\|^2_2 + \|A_t-S_t\|^2_2 + TV(A_s) + TV(A_t) \\
TV(A)&=\sum_{i,j}[A(i,j)-A(i-1,j)]^2 + [A(i,j)-A(i,j-1)]^2.
\end{aligned}
\label{equ:att}
\end{equation}
For the generator, the full objective function is shown as follows, and
$\lambda_p, \lambda_f$ and $\lambda_a$ are the weights of perceptual, face identity and attention losses, respectively.
\begin{equation}
\begin{aligned}
L^G = \lambda_p L_p + \lambda_f L_f + \lambda_a L_a + L^G_{adv}.
\end{aligned}
\label{equ:l_g}
\end{equation}
For discriminator, the full objective function is
\begin{equation}
\small
\begin{aligned}
L^D &= \sum[D_{Global}(\hat{I}_t, C_t) + 1]^2 + \sum [D_{Global}(I_r, C_t) - 1]^2 \\
&+ \sum[D_{Body}(\hat{I}^b_t), C^b_t) + 1]^2 + \sum[D_{Body}(I^b_r, C^b_t) - 1]^2 \\
&+ \sum[D_{Head}(\hat{I}^h_t, C^h_t) + 1]^2 + \sum[D_{Head}(I^h_r, C^h_t) - 1]^2.
\end{aligned}
\label{equ:l_d}
\end{equation}

\begin{figure*}[t]
	\begin{center}
		\includegraphics[width=\linewidth]{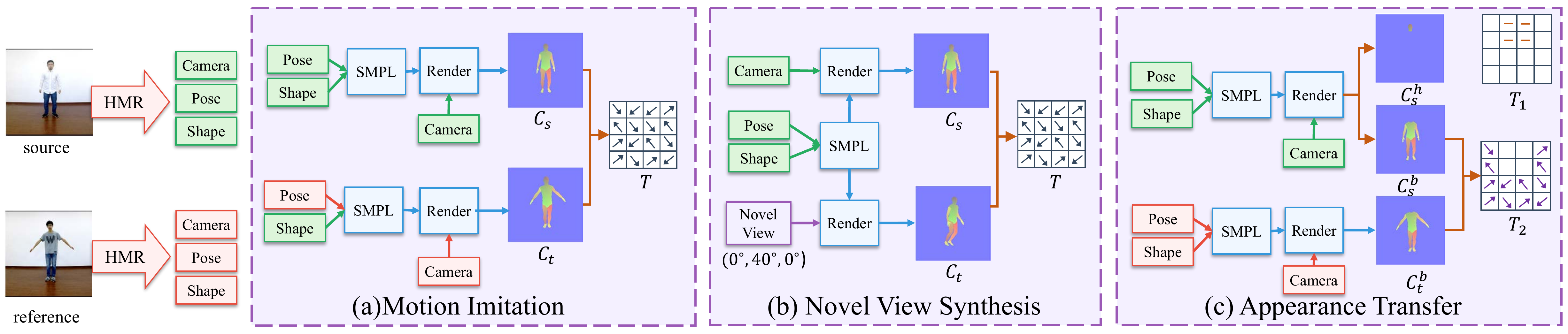}
	\end{center}
	\caption{Illustration of calculating the transformation flows of different tasks during the testing phase. The left is the disentangled body parameters by the Body Recovery module of both source and reference images. The right is the different implementations to calculate the transformation flow in different tasks.}
	\label{fig:inference}
\end{figure*}

\subsection{One/Few-shot Personalization by Fine-tunning}
Though we can train our model on a large dataset, to a certain degree, with diverse people and clothes, however, such a generator is still hard to be well-generalized to the inputs out of the domain of training set. After all, it is infeasible to build a universal dataset and generator to handle the diverse face identities, styles of clothes, and backgrounds.  To improve the generalization, inspired by the SinGAN~\cite{Shaham_2019_ICCV} and the Meta-learning~\cite{Zakharov_2019_ICCV, Wang_2019_NeurIPS, Lee_2019_ICLR_MetaPix, Finn_2017_MAML_ICML}, we apply the one/few-shot adversarial learning to push the network to focus on each individual by several steps of fast personal adaptation. In real application scenarios, the user might only provide a little number ($s_n$) of their photos with different views or poses, and in an extreme case, there is only one image accessible.  In this paper, we focus on the setting where there are no more than eight images ($s_n \le 8$)~\cite{Zakharov_2019_ICCV} available in the testing phase.

Specifically, we first train our model, including a generator and a discriminator, on a combined large dataset, and consequently obtain the generator's pre-trained parameters, $ \theta^M_G $, and the discriminator's pre-trained parameters, $\theta^M_D$. Then, for each specific person $P_i$ with $s_n$ images, we learn the person-specific generator $\theta^{P_i}_G$ and discriminator $\theta^{P_i}_D$ from the $s_n$ images by fine-tuning the pre-trained model. This process is called one/few-shot personalization. To further push the generator from the pre-trained  $\theta^M_{G}$ to the person-specific $\theta^{P_i}_{G}$, we discard the pre-trained parameters of the discriminator $\theta^M_D$, and we train the person-specific discriminator $\theta^{Pi}_D$ from scratch. The overall loss functions in the personalization phase are similar to that in the training phase, except for the adversarial loss. Since there are only a few images ($s_n \le 8$), to avoid overfitting and reduce the time consumption of each iteration in personalization, we only use the global discriminator.

\subsection{Inference}
\label{bookmark:inference}
After we conduct personalization, the person-specific generator can be applied to all three tasks. The difference lies in the transformation flow computation, due to the different conditions of various tasks. The remaining modules, Body Mesh Recovery and Liquid Warping GAN (Attentional Liquid Warping GAN) are all the same. The followings are the details of each task of the Flow Composition module in the testing phase.

\textbf{Motion Imitation.} We firstly copy the value of pose parameters of the reference $\theta_r$ into that of the source and get the synthetic parameters of SMPL, as well as the 3D mesh, $M_t = M(\theta_r, \beta_s)$. Next, we render a correspondence map of the source mesh $M_s$ and that of the synthetic mesh $M_t$ under a camera view $K_s$. Here, we denote the source and the synthetic correspondence map as $C_s$ and $C_t$, respectively. Then, we project the source vertices into the 2D image space by a weak-perspective camera, $v_s=\pi(V_s, K_s)$. Here, $\pi$ is the weak-perspective projective function. Next, we calculate the barycentric coordinates of each mesh face and have $f_s \in \mathbb{R}^{N_f \times 2} $. Finally, we calculate the transformation flow $T\in\mathbb{R}^{H\times W\times 2}$ by matching the correspondences between the source correspondence map with its mesh face coordinates $f_s$ and the synthetic correspondence map. It is shown in Fig.~\ref{fig:inference} (a).

\textbf{Novel View Synthesis.} Given a new camera view, in terms of a rotation $R$ and a translation $t$. We firstly calculate the 3D mesh under the novel view, $M_t = M_sR + t$. The consequential operations are similar to that of motion imitation. We render a correspondence map of the source mesh $M_s$ and that of the novel mesh $M_t$ under a weak-perspective camera $K_s$ and calculate the transformation flow $T\in\mathbb{R}^{H\times W\times 2}$, as depicted in Fig.~\ref{fig:inference} (b).

\textbf{Appearance Transfer.} We need to ``copy'' the clothes on the body from the reference image while keeping the head (face, eye, hair and so on) identity of the source. We split the transformation flow $T$ into two sub-transformation flows, source flow $T_1$ and referent flow $T_2$. We denote the head mesh as $M^{h} = (V^{h}, F^{h})$ and the body mesh as $M^{b} = (V^{b}, F^{b})$. Here, $M = M^{h} \cup M^{b}$.  For $T_1$, We firstly project the head mesh $M^{h}_s$ of source into the image space and thereby obtain the silhouettes, $S^{h}_s$. Then, we create a mesh grid, $G\in\mathbb{R}^{H\times W\times 2}$. Next, we mask $G$ by $S^{h}$ and derive $T_1 = G \odot S^{h}$. Here, $\odot$ represents an element-wise multiplication. For $T_2$, it is similar to that in motion imitation. We render the correspondence map of the source body $M^{b}_s$ and that of the reference $M^{b}_t$, denoted as $C^{b}_s$ and $C^{b}_t$, respectively. Finally, we calculate the transformation flow $T_2$ based on the correspondences between $C^{b}_s$ and $C^{b}_t$. We illustrate it in Fig.~\ref{fig:inference} (c).

\begin{figure}[h]
	\centering
	\includegraphics[width=\linewidth]{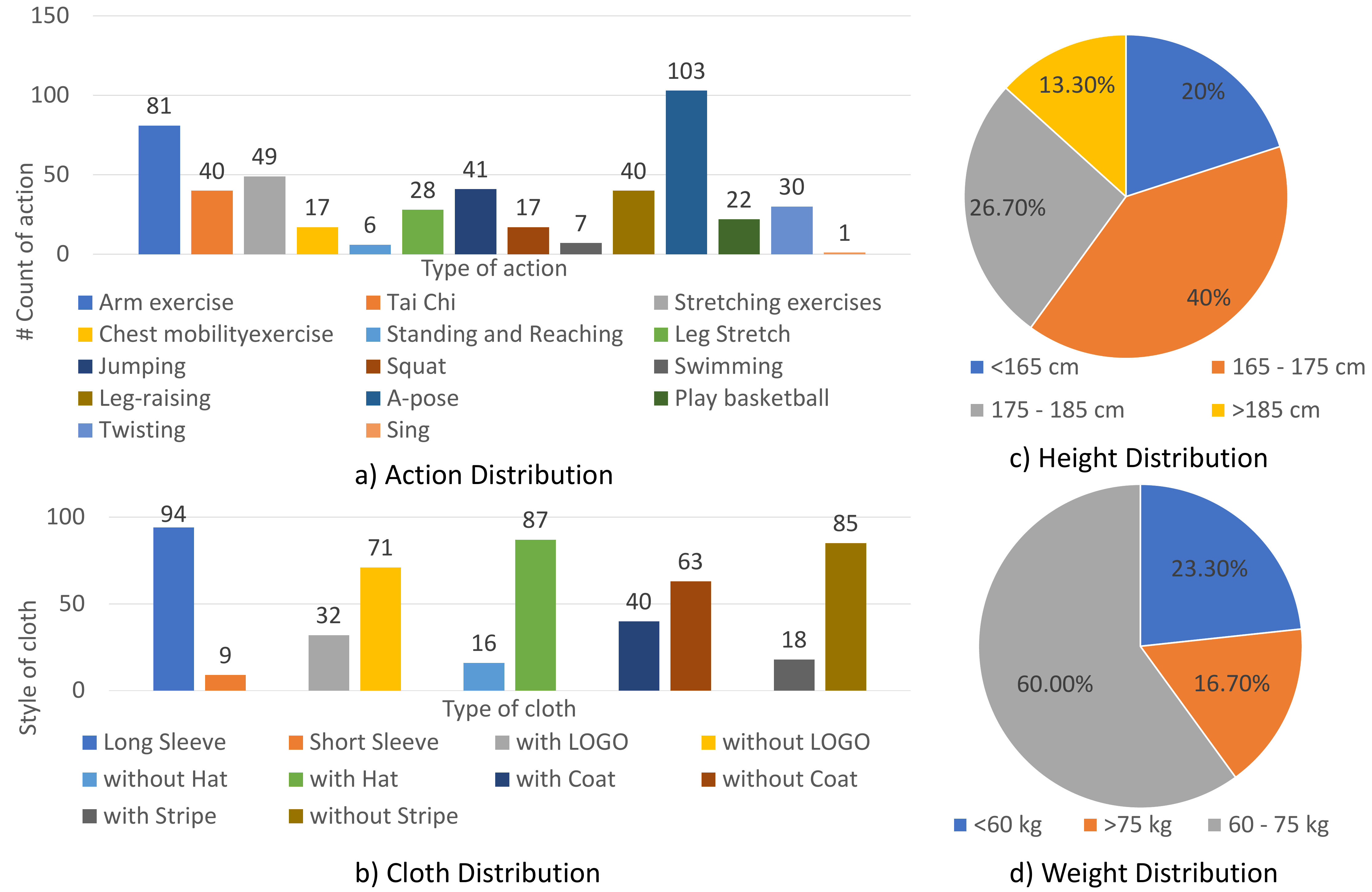}
	\caption{The statistic information of iPER dataset, including the action, clothes, height and weight distribution of the actors.}
	\label{fig:iPER_info} 
\end{figure}

\section{Experiments}
\subsection{Dataset}
\begin{figure}[h]
	\centering
	\includegraphics[width=0.9\linewidth]{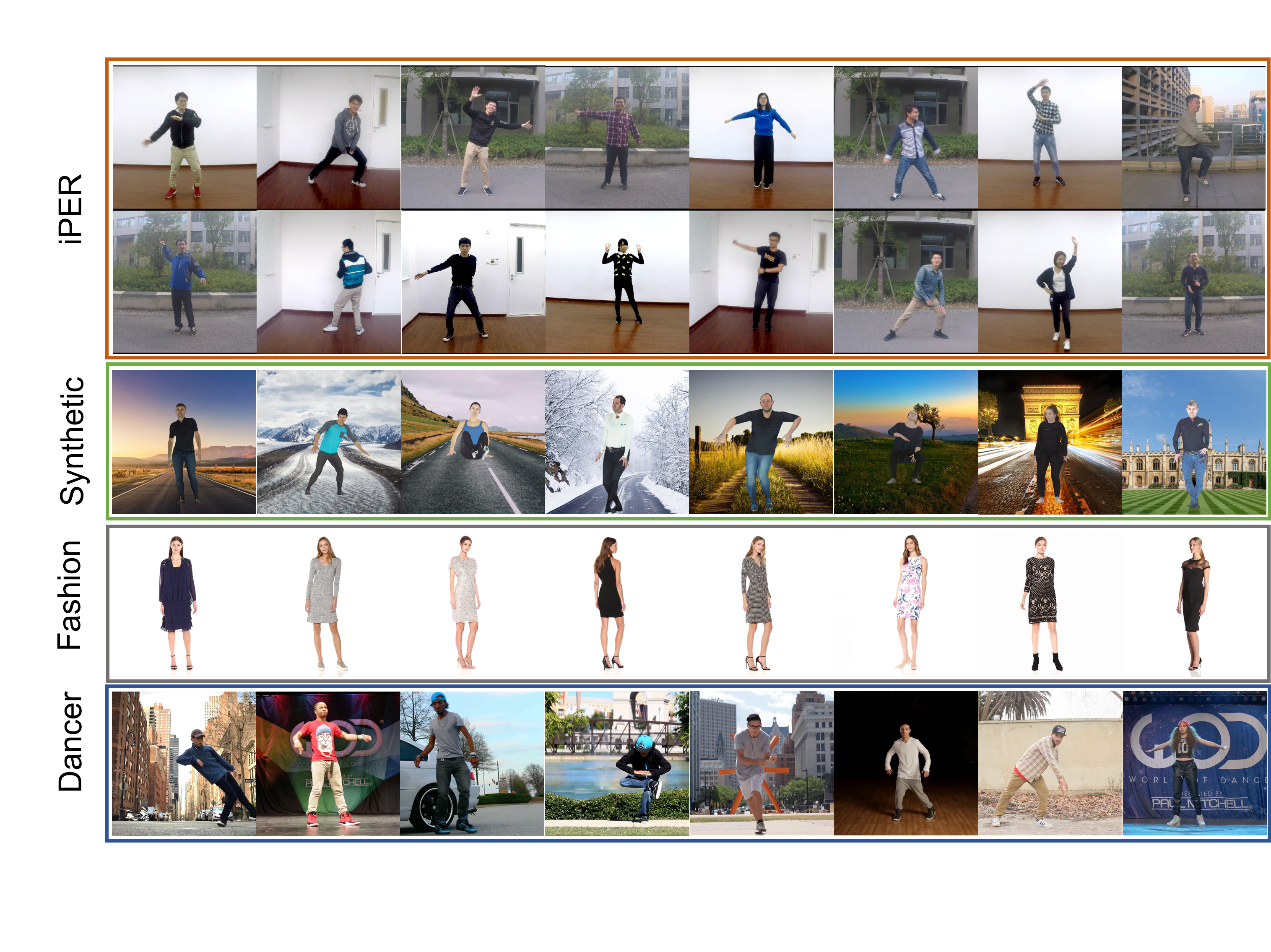}
	\caption{The samples of four datasets. The first two rows are the samples from iPER dataset. The third row is the samples from the MotionSynthetic dataset and the fourth row is that from FashionVideo dataset. The last row is the samples from Youtube-Dancer-18 dataset.}
	\label{fig:dataset_samples} 
\end{figure}

\textbf{iPER.} To evaluate the performance of our proposed method of motion imitation, appearance transfer, and novel view synthesis, we build a new dataset with diverse styles of clothes in videos, named Impersonator (iPER) dataset. There are 30 subjects of different conditions of shape, height, and gender. Each subject wears different clothes and performs an A-pose video and a video with random actions. There are 103 clothes in total.
The whole dataset contains 206 video sequences with 241,564 frames. We split it into training/testing set at a ratio of 8:2 according to the different clothes. All the clothes and 29\% of the actors in the testing set do not appear in the training set. We illustrate the details of the iPER dataset in classes of actions, styles of clothes, weight, and height distributions of actors in Fig.~\ref{fig:iPER_info}. 
We show some samples in the first two rows of Fig.~\ref{fig:dataset_samples}.

\textbf{MotionSynthetic.} We also make up a synthetic dataset, named MotionSynthetic, for the convenience of evaluation, especially for human appearance transfer and novel view synthesis, because we can synthesize the ground truth images with different views and wearing garments by the modification of meshes. This dataset borrows 24 human meshes from people snapshot~\cite{Alldieck_2018_CVPR} and 96 human meshes from MultiGarments~\cite{Bhatnagar_2019_ICCV}; thus, 120 meshes in total. All of these meshes with UV texture images have been registered in SMPL~\cite{SMPL:2015}.  For each mesh, we choose a pose sequence from Mixamo and a background image from the Internet. Based on these materials (mesh, UV image, pose sequence, and background image), we render the synthetic images by NMR~\cite{cvprKatoUH18}, resulting in 39,529 frames in total. We split it into training/testing set at a ratio of 8:2 according to the different meshes and illustrate some synthetic images in the 3rd rows of Fig.~\ref{fig:dataset_samples}.

\textbf{FashionVideo.} It contains 500 training and 100 testing videos with a single female model wearing fashionable clothes~\cite{fashionvideo_bmvc_2019}. Each video has around 350 frames. The clothes and textures are diverse, while there are few types of gestures, with only a few standard poses for the models. Also, this dataset lacks diversity in background, and all the backgrounds are black. We display some samples in the 4th row of Fig~\ref{fig:dataset_samples}.

\textbf{Youtube-Dancer-18.} To further validate the effectiveness and generalization of our method, we evaluate our method on the in-the-wild internet videos, Youtube-Dancer-18~\cite{Lee_2019_ICLR_MetaPix}.  It consists of 18 videos, with people dancing, downloaded from Youtube, and each of them lasts from 4 to 12 minutes. We follow the setting with MetaPix~\cite{Lee_2019_ICLR_MetaPix} that we sample frames with 30 FPS and only use $s_n \le 8$ frames from training sequences for personalization and then apply the evaluation on the testing sequences. Some samples are shown at the bottom of Fig.~\ref{fig:dataset_samples}. It needs to be mentioned that we do not train the model in this dataset. We only sample $s_n$ frames for personalization and directly test on this dataset to evaluate the generalization over all methods.

\subsection{Implementation Details}
We train our Attentional Liquid Warping GAN on a combined dataset consisting of the iPER,  MotionSynthetic, and FashionVideo dataset and perform evaluations among these three datasets. To evaluate our methods' generalization, we also perform tests on an additional Youtube-Dancer-18 dataset without training on it. We crop all images based on the bounding box of the human body, rescale the cropped images with keeping the original ratio of height and width, and then pad them into a $512\times512$ resolution. We normalize the color space of all images to [-1, 1]. In our experiments, including the training and personalization phase, we use the Adam~\cite{Kingma2015AdamAM} based Stochastic Gradient Descent optimizer for both generators and discriminators. $\lambda_p, \lambda_f$ and $\lambda_a$ are 10.0, 5.0 and 2.5, respectively. 

i): In the training phase, we randomly sample $s_n + 1$ images from each video and set the first $s_n$ ones as the source images $\{I_{s_1}, ..., I_{s_n} \}$, and the other one as the reference $I_r$. We fix $s_n = 2$ and the mini-batch size to be 2. There are two training epochs. We fix the first quarter training session with a learning rate as 0.0001 and gradually decrease it to 0.00001 in the end.

ii): In the personalization and testing phase, $s_n$ could be flexible, and because of the memory limitation of the GPU devices, in our experiments, we set $s_n \in \{1, 2, 4, 8\}$. Besides, $I_r$ lies in the set of source images $\{I_{s_1}, ..., I_{s_n} \}$. We fix the learning rate as 0.0001 and take $T=100$ steps for personalization. 

\subsection{Results of Human Motion Imitation}
\begin{figure*}[t]
\centering
\includegraphics[width=\linewidth]{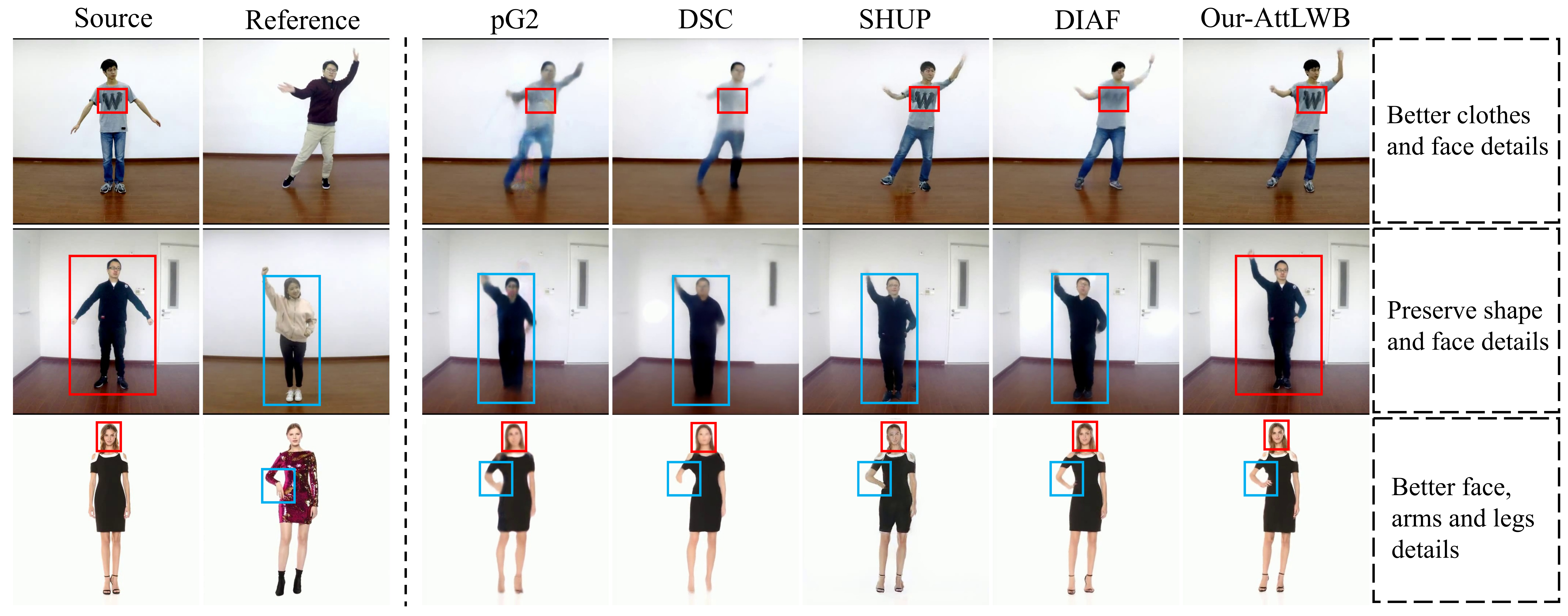}
\caption{Comparison of our method with others of motion imitation on the iPER and FashionVideo dataset (zoom-in for the best of view). All results are in $512 \times 512$ resolution. 2D pose-guided methods~pG2~\cite{pG2017nips}, DSC~\cite{DSC2018} SHUP~\cite{posewarp2018} and DIAF cannot preserve the clothes details, face identity and shape consistency of source images. We highlight the details by red and blue rectangles.}
\label{fig:comparison} 
\end{figure*}

\begin{figure*}[t]
\centering
\includegraphics[width=\linewidth]{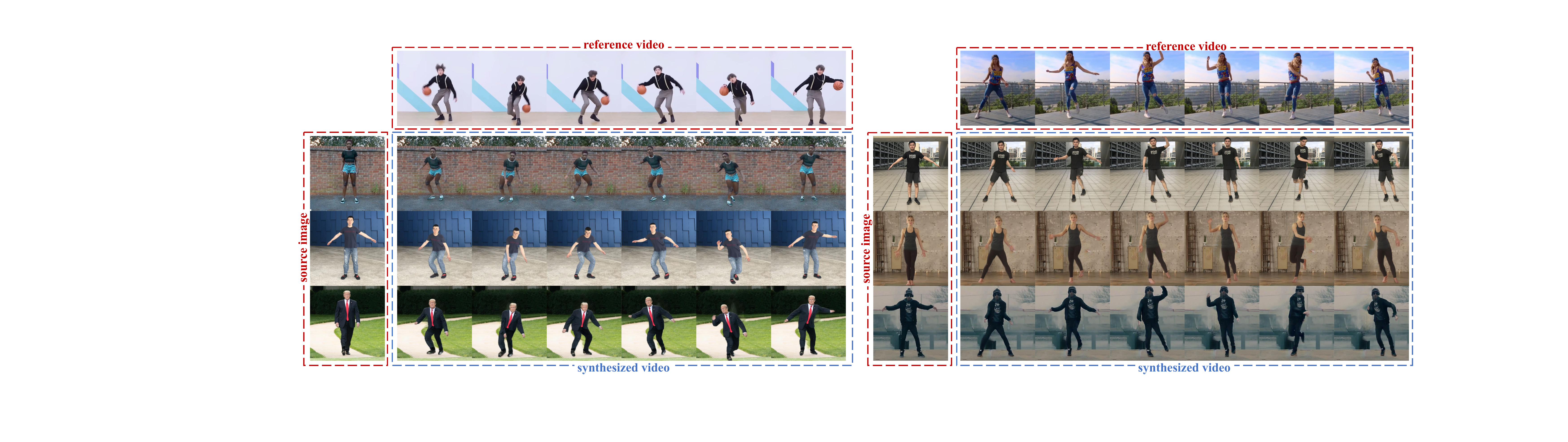}
\caption{Examples of motion imitation from our proposed methods (zoom-in for the best of view). All results are in $512 \times 512$ resolution. Our method could produce high-fidelity images that preserve the face identity, shape consistency and clothes details of source. We recommend accessing the supplementary material for more results in videos.}
\label{fig:imitation} 
\end{figure*}

\textbf{Evaluation Metrics.} We propose an evaluation protocol of the testing set of the iPER, MotionSynthetic, FashionVideo, and Youtube-Dancer-18 datasets, and it can indicate the performance of different methods in terms of different aspects. The details are listed in followings:

1): In each video with actor $P_i$, $\{I^{P_i}_1, ..., I^{P_i}_t, ..., I^{P_i}_L\}$, we select eight images as candidate images with different views, such as frontal, sideways or back. Here, $L$ is the number of frames.

2): We choose $s_n \le 8$ images as sources, $\{I^{P_i}_{s_1},...,I^{P_i}_{s_n}\}$, from the eight candidate images for personalization. For a fair comparison with other methods~\cite{pG2017nips,posewarp2018,DSC2018,Li_2019_CVPR,Zhu_2019_CVPR}, which only use a single source image, we separately report the results on $s_n = 1$ (one-shot setting) and $2\le s_n \le 8$(few-shot setting). 

3): After personalization, we perform self-imitation that each actor $P_i$ imitates actions from images of themselves, with $I^{P_i}_t$ as the reference image. We denote $\hat{I}^{P_i\to P_i}_t$ as the synthesized image referring to $I^{P_i}_t$. As for criterion, we use PSNR, SSIM~\cite{ssimWangBSS04}, Learned Perceptual Similarity (LPIPS)~\cite{zhang2018perceptual}, Body-CS and Face-CS to measure the similarities between $\hat{I}^{P_i\to P_i}_t$ and $I^{P_i}_t$.

\textbf{Body-Cosine-Similarity (Body-CS)}: is the distance between the cropped person region of the 
synthesized image and that of the ground-truth image. In particular, it firstly uses a YOLOv3~\cite{yolo2016} detector to get the person bounding box of the synthesized and ground-truth image. Then, we crop the person patches according to the bounding boxes. Finally, we use a pre-trained Person re-identification (ReID) model, OS-Net~\cite{Zhou_2019_ICCV}, to get the embedding features of the cropped person patches, and then we normalize the features and calculate the cosine similarity between the features to acquire the Body-CS.

\textbf{Face-Cosine-Similarity (Face-CS)}: similar to Body-CS, it is the distance between the cropped face region of the synthesized image and that of the ground-truth image. Specifically, we firstly use an MTCNN~\cite{mtcnn2016} face detector to get the face bounding boxes of the synthesized and ground-truth images. Then, we crop the face regions according to the bounding boxes. Finally, we uses a pre-trained face recognition model~\cite{InceptionResnetV1},
to get the embedding features of the cropped face patches, and then we normalize the features and calculate the cosine similarity between the normalized features to obtain the Face-CS.

4): We also conduct cross-imitation that an actor $P_i$ imitates actions from others, such as $P_j$. We denote $\{\hat{I}^{P_i\to P_j}_1, ..., \hat{I}^{P_i\to P_j}_L\}$ as a sequence of synthesized images referring to $\{I^{P_j}_1, ..., I^{P_j}_L\}$ and $\{I^{P_i}_{s_1},...,I^{P_i}_{s_n}\}$ as the sequence of real images. Since there is no ground-truth of synthesized images for the similarities metrics as mentioned above, here, we use a Fr\'{e}chet Inception Distance (FID)~\cite{fid2017} to measure perceptual realism. It calculates the distance between the set of synthesized images and that of real images. We further propose the Fr\'{e}chet Distance of a pre-trained ReID model, OS-Net~\cite{Zhou_2019_ICCV}, namely Body-FD and that of a face recognition model, namely Face-FD. We also collect $L$ consecutive frames from the actor $P^i$, denoted as $\{I^{P_i}_1, ..., I^{P_i}_L\}$, then calculate the Body-CS and Face-CS as aforementioned.

\begin{table*}[t]
\centering
\caption{\textbf{One-shot} average results for human motion imitation of different methods on the iPER, MotionSynthetic and FashionVidieo dataset.
	$\uparrow$ means the larger the better, and $\downarrow$ is on the contrary. A higher SSIM may not mean a better quality of an image~\cite{zhang2018perceptual}.}
\begin{tabular}{c|c|c|c|c|c|c|c|c|c|c}
	\hline
	& \multicolumn{5}{c|}{Self-Imitation}              & \multicolumn{5}{c}{Cross-Imitation}                   \\ \cline{2-11} 
	& PSNR$\uparrow$   & SSIM$\uparrow$   & LPIPS$\downarrow$  & Body-CS$\uparrow$ & Face-CS$\uparrow$ & Face-CS$\uparrow$  & Face-FD$\downarrow$  & Body-CS$\uparrow$ & Body-FD$\downarrow$ & FID$\downarrow$      \\ \hline
	PG2~\cite{pG2017nips}           & 23.699          & 0.876         & 0.130           & 0.744         & 0.085           & 0.148         & 429.142         & 0.709         & 240.429         & 119.378         \\
	SHUP~\cite{posewarp2018}         & 23.979          & \textbf{0.881} & 0.080           & 0.855          & 0.288           & 0.297          & 243.599         & 0.820          & 80.973          & 51.823          \\
	DSC~\cite{DSC2018}            & 20.782          & 0.732          & 0.331          & 0.695          & 0.139          & 0.204          & 407.070         & 0.673          & 273.103         & 150.082         \\
	DIAF~\cite{Li_2019_CVPR}           & 22.753          & 0.829          & 0.108          & 0.851          & 0.390          & 0.364          & 166.560         & 0.808          & 102.807         & 63.528          \\
	PATB~\cite{Zhu_2019_CVPR}                 & 20.387          & 0.798          & 0.169          & 0.738          & 0.129          & 0.363          & 218.333        & 0.731          & 259.135        & 136.911         \\
	\textbf{Our-LWB}  & {23.932} & {0.843} & {0.089} & {0.901} & {0.560} & {0.538} & {99.258} & {0.862} & {48.619} & {32.370} \\
	\textbf{Our-AttLWB}  & \textbf{24.513} & 0.856          & \textbf{0.074} & \textbf{0.911} & \textbf{0.591} & \textbf{0.564} & \textbf{73.217} & \textbf{0.869} & \textbf{44.022} & \textbf{30.503}
	\\ \hline
\end{tabular}
\label{table:one_shot}
\end{table*}

\begin{table*}[t]
\centering
\caption{\textbf{Few-shot} results for human motion imitation of different methods on the Youtube-Dancer-18 dataset. The number of source images $s_n$ is $2$. $\uparrow$ means the larger the better, and $\downarrow$ represents the smaller the better. }
\begin{tabular}{c|c|c|c|c|c|c|c|c|c|c}
	\hline
	& \multicolumn{5}{c|}{Self-Imitation}              & \multicolumn{5}{c}{Cross-Imitation}                   \\ \cline{2-11} 
	& PSNR$\uparrow$    & SSIM$\uparrow$   & LPIPS$\downarrow$   & Body-CS$\uparrow$ & Face-CS$\uparrow$   & Face-CS$\uparrow$  & Face-FD$\downarrow$   & Body-CS$\uparrow$ & Body-FD$\downarrow$ & FID$\downarrow$      \\ \hline
	pix2pixHD~\cite{Wang_2018_CVPR}         & 11.134          & 0.196          & 0.633           & 0.616          & 0.106           & 0.136          & 221.661         & 0.565          & 266.552         & 175.574         \\
	SPADE~\cite{park2019SPADE}      & 8.984           & 0.120          & 0.780           & 0.535          & 0.106           & 0.131          & 294.672         & 0.513          & 431.670         & 304.698         \\
	MetaPix Pix2PixHD~\cite{Lee_2019_ICLR_MetaPix} & 14.052          & 0.385          & 0.550           & 0.549          & 0.134           & 0.187          & 277.555         & 0.523          & 441.495         & 257.457         \\
	MetaPix SHUP~\cite{Lee_2019_ICLR_MetaPix}                   & 18.857        & \textbf{0.649} & 0.269            & 0.765          & 0.234           & 0.191          & 185.363        & 0.693          & 160.485         & 83.501          \\
	\textbf{Our-LWB}                 & {19.485}  & {0.642}  & {0.245} & {0.830}  & {0.413}   & {0.355} & {96.280}   & {0.738}     & {102.075}   & {70.743}
	\\
	\textbf{Our-AttLWB}                   & \textbf{19.691} & \textbf{0.649} & \textbf{0.232} & \textbf{0.831} & \textbf{0.437} & \textbf{0.380} & \textbf{82.053} & \textbf{0.743} & \textbf{99.575} & \textbf{65.454}
	\\ \hline
\end{tabular}
\label{table:few_shot}
\end{table*}

\textbf{Quantitative Comparison with Other Methods under One-shot Setting.} We compare the performance of our method with that of existing methods, including PG2~\cite{pG2017nips}, SHUP~\cite{posewarp2018}, DSC~\cite{DSC2018}, DIAF~\cite{Li_2019_CVPR} and PATB~\cite{Zhu_2019_CVPR}. We train all these methods on a combined dataset with the iPER, MotionSynthetic, and FashionVideo dataset and apply the evaluation protocol with the one-shot setting mentioned above to these methods. We report the results in Table~\ref{table:one_shot}, and our method outperforms others on all the metrics except SSIM, for which a higher numerical value does not necessarily mean a better quality of an image as reported in~\cite{zhang2018perceptual}.

\textbf{Quantitative Comparison with Other Methods under Few-shot Setting.} We compare the performance of our method with pix2pixHD~\cite{Wang_2018_CVPR}, SPADE~\cite{park2019SPADE}, MetaPix pix2pixHD and MetaPix SHUP~\cite{Lee_2019_ICLR_MetaPix} under this setting. Here, we report the results on the Youtube-Dancer-18 dataset with the number of source images $s_n$ being $2$ in Table~\ref{table:few_shot} and our method outperforms others.

\textbf{Qualitative Comparison.} Besides, we also analyze the generated images and make comparisons between ours and the above methods. From Fig.~\ref{fig:comparison}, we find that 1) the above methods that use 2D pose-guided inputs change the body shape of the source. For example, in the $2^{nd}$ row of Fig.~\ref{fig:comparison}, the scenario is a tall person imitating motion from a short person, and baseline methods change the height of the source body. However, our method is capable of keeping the body shape unchanged because our method disentangles the pose and the personalized shape of each actor.
2) In the light of our proposed AttLWB (LWB) and face identity loss, our method is more powerful in terms of preserving source identities, such as the face identity and cloth details of source than other methods, as shown in the $1^{st}$ and $2^{nd}$ row of Fig.~\ref{fig:comparison}. 
3) Our method also produces high-fidelity images in the cross-imitation setting (imitating actions from others), which we illustrate in Fig.~\ref{fig:imitation}. As we can see in Fig.~\ref{fig:imitation}, the face identity, and clothes details, in terms of texture color and style, are preserved well. It shows that our method can achieve decent results in cross imitation even when the reference image comes from the Internet, which is out of the domain of our training dataset.

\subsection{Results of Human Appearance Transfer}
\begin{figure*}[t]
	\centering
	\includegraphics[width=0.9\linewidth]{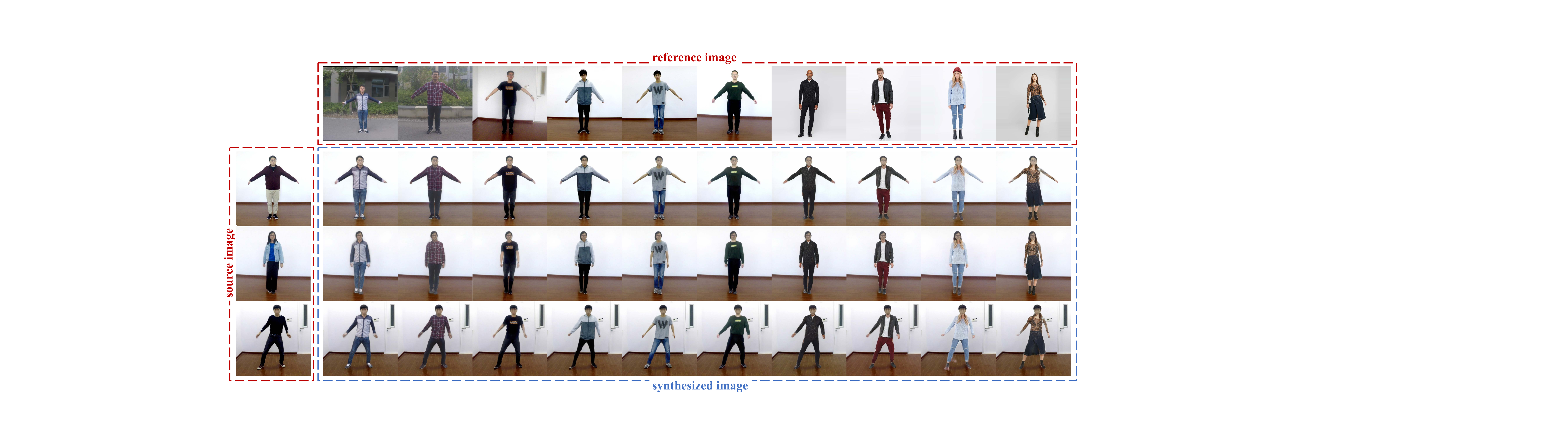}
	\caption{Examples of our proposed AttLWB of human appearance transfer in the testing set of iPER (zoom-in for the best of view). All results are in $512 \times 512$ resolution. Our method could produce high-fidelity and decent images that preserve the face identity and shape consistency of the source image and keep the clothes details of reference image. }
	\label{fig:exam_app} 
\end{figure*}

\begin{figure*}[t]
	\centering
	\includegraphics[width=0.9\linewidth]{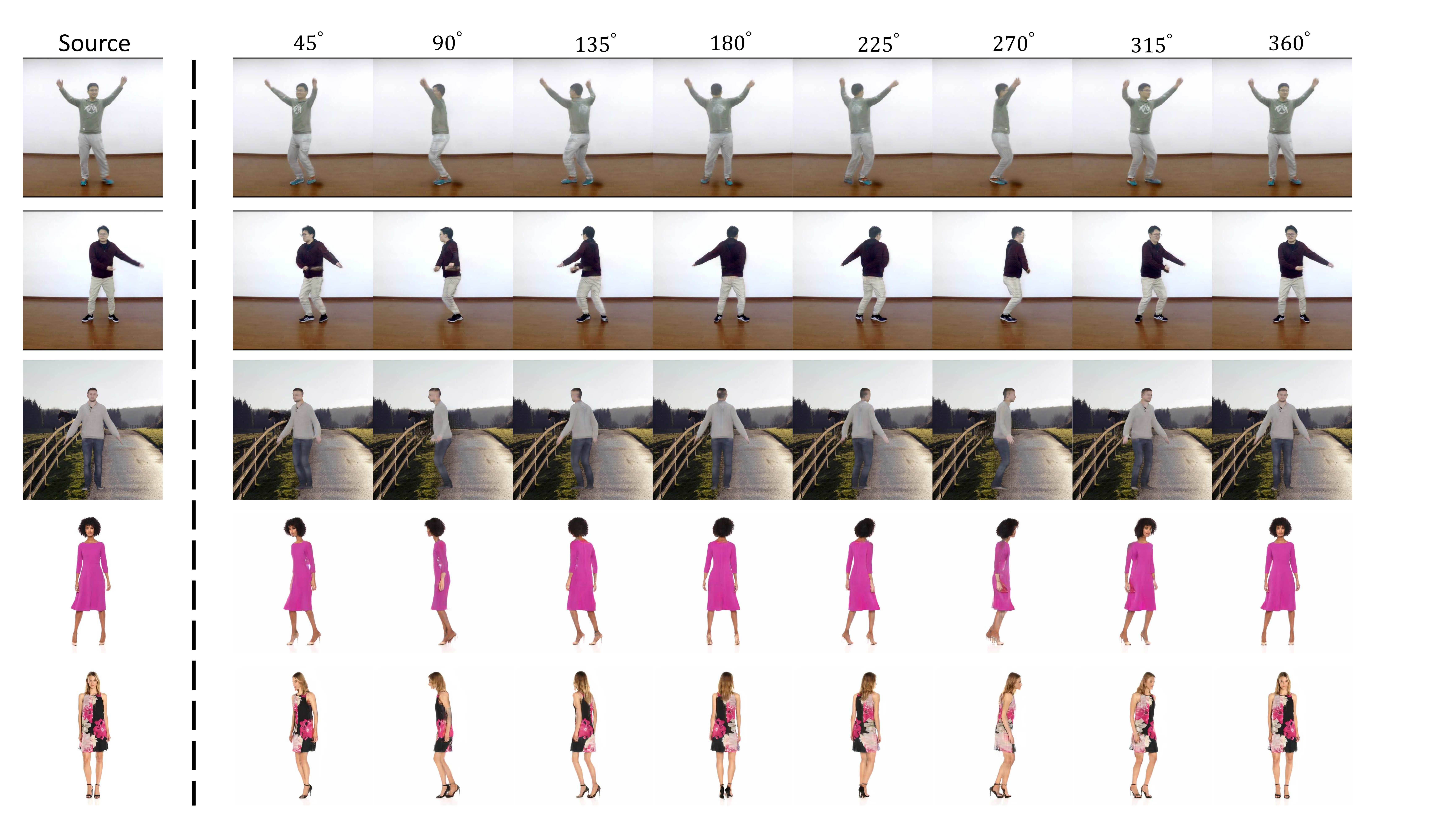}
	\caption{Examples of our proposed AttLWB of human novel view synthesis. It is capable of preserving the source information, in terms of face identity and logo details of cloths,  even the person wearing the long dress with fluffy hair.}
	\label{fig:novel} 
\end{figure*}


It is worth emphasizing that once the model has been trained, it can directly be applied in three tasks, including motion imitation, appearance transfer, and novel view synthesis. We conduct the experiments on the iPER dataset.

\textbf{Evaluation Metrics.} In the iPER dataset, subjects might wear different clothes, and we sample the same person's images with different clothes as the source and the reference image. We use aforementioned PSNR, SSIM~\cite{ssimWangBSS04}, LPIPS~\cite{zhang2018perceptual}, Body-CS and Face-CS as the metrics.

\textbf{Quantitative Results.} We report the results of our methods with LWB and AttWLB on the iPER dataset in Table~\ref{table:appearance_transfer}. The results show that Attentional Liquid Warping Block (AttLWB) is slightly better than the LWB.
\begin{table}[h]
\centering
\caption{Results for human appearance transfer of our LWB and AttLWB, on the iPER dataset. Here, we report the PSNR, SSIM, LPIPS, Body-CS and Face-CS. $\uparrow$ means the larger the better. A higher SSIM may not mean a better quality of an image~\cite{zhang2018perceptual}.}
\begin{tabularx}{\linewidth}{p{1.7cm}<{\centering}|X|X<{\centering}|X|p{1.22cm}<{\centering}|p{1.12cm}<{\centering}}
	\hline
	&PSRN$\uparrow$ & SSIM$\uparrow$ & LPIPS$\downarrow$ & Body-CS$\uparrow$ & Face-CS$\uparrow$ \\ \hline
	\textbf{Our-LWB}     &17.707 & \textbf{0.734} & 0.225 & 0.891 & 0.642 \\
	\textbf{Our-AttLWB}  &\textbf{17.783} & 0.726 & \textbf{0.220} & \textbf{0.896} & \textbf{0.706} \\ \hline
\end{tabularx}
\label{table:appearance_transfer}
\end{table}

\textbf{Qualitative Results.} We randomly pick some examples displayed in Fig.~\ref{fig:exam_app}. The face identity and clothes details, in terms of texture, color, and style, are preserved well by our method. It demonstrates that our method can achieve decent results in appearance transfer, even when the reference image comes from the Internet and is out of the domain of the iPER dataset, such as the last five columns in Fig.~\ref{fig:exam_app}.

\subsection{Results of Human Novel View Synthesis}
\textbf{Evaluation Metrics.} As for data in the iPER dataset, we have videos containing different views of a certain subject performing A-pose, and in the MotionSynthetic dataset, we render A-pose images with 3D meshes from different viewpoints. Thus, we obtain images of the same person in different views. For evaluation, we use PSNR, SSIM~\cite{ssimWangBSS04} and LPIPS~\cite{zhang2018perceptual} as the metrics.

\begin{table}[h]
\centering
\caption{Results for human novel view synthesis of different methods, including AppFlow~\cite{ZhouTSME16}, MV2NV~\cite{sun2018multiview}, ours LWB and AttLWB, on iPER and MotionSynthetic dataset. Here, we report the PSNR, SSIM and LPIPS~\cite{zhang2018perceptual}. $\uparrow$ means the larger the better. A higher SSIM may not mean a better quality of an image~\cite{zhang2018perceptual}.}
\begin{tabularx}{8.7cm}{p{0.9cm}<{\centering}|X<{\centering}|X<{\centering}|X<{\centering}|X<{\centering}|X<{\centering}|X<{\centering}}
	\hline
	&
	\multicolumn{3}{c|}{iPER} &
	\multicolumn{3}{c}{MotionSynthetic}
	\\ \cline{2-7} 
	&
	PSNR$\uparrow$ &
	SSIM$\uparrow$ &
	LPIPS$\downarrow$ &
	PSNR$\uparrow$ &
	SSIM$\uparrow$ &
	LPIPS$\downarrow$
	\\ \hline
	AppFlow         & 23.342                    & 0.849                     & 0.133                     & 25.575                    & 0.896                     & 0.083                       \\
	MV2NV           & 24.950                    & \textbf{0.883}            & 0.125                     & 25.951                    & 0.837                     & 0.097                       \\
	\textbf{LWB}              & 
	24.518         			  & 
	0.862          			  & 
	0.090                     &             25.055                    & 
	0.779                     & 
	0.106                       \\
	\textbf{AttLWB} & \textbf{25.246}           & 0.867                     & \textbf{0.078}               & \textbf{28.625}           & \textbf{0.934}            & \textbf{0.037}              \\ \hline
	
\end{tabularx}
\label{table:novel_view}
\end{table}

\textbf{Quantitative Results.} In Table~\ref{table:novel_view}, we report the results of our methods AttLWB and that of other state-of-the-art methods, including AppFlow~\cite{ZhouTSME16} and MV2NV~\cite{sun2018multiview}, on the iPER and MotionSynthetic datasets based on the above evaluation metrics. The results show that our method outperforms other methods.

\textbf{Qualitative Results.} We randomly sample source images from the testing set of the iPER dataset and change the views from $30^{\circ}$ to $330^{\circ}$. The results are illustrated in Fig.~\ref{fig:novel}. Our method is capable of predicting reasonable content of invisible parts when switching to other views and keep the source information, in terms of face identity and clothes details, even in the self-occlusion case, such as the middle and bottom rows in Fig.~\ref{fig:novel}. Through Fig.~\ref{fig:novel}, we can see that 1) even when the subjects have large motion deformation, such as the case in the $1^{st}$ row of Fig.~\ref{fig:novel}, results of our method can keep the logo details of clothes. 2) The $2^{nd}$ row shows the results when the subjects have self-occlusion. 3) Our method can also handle cases with complex background as the $3^{rd}$ row in Fig.~\ref{fig:novel} shows. 4) The $4^{th}$ row of Fig.~\ref{fig:novel} shows cases in which subjects wear a long dress and have fluffy hair. 5) The $5^{th}$ row of Fig.~\ref{fig:novel} is the case with complex clothes texture.

\section{Ablation Studies and Analysis}
In this section, we perform experiments to analyze the impacts of factors in our system, including with/without personalization, ablation studies of different loss functions and the comparison of our proposed LWB or AttLWB with other warping strategies, such as input concatenation, texture warping and feature warping. We further report the running time and analyze the failure cases.

\subsection{Impact of Personalization}
\begin{figure*}[t]
	\centering
	\includegraphics[width=0.9\linewidth]{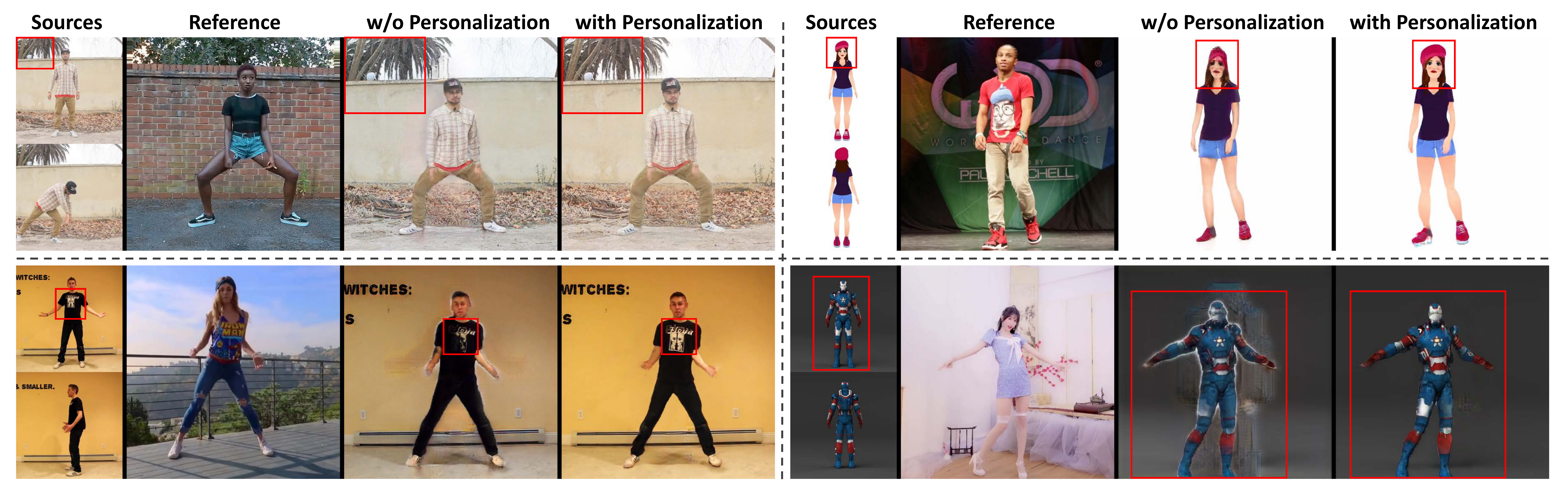}
	\caption{Comparison of our proposed AttLWB with and without personalization (zoom-in for the best of view). The two roles in the left column are the source images from the Youtube-Dancer-18 dataset, and that in the right column are cartoon images from the Internet. From the top left segment, we can see that our method could preserve the color style of the source background with personalization. From the bottom-left segment, we find that our method without personalization might lose the details of the logo structure in the source images, while our method with personalization could preserve the logo details. The roles in the right column demonstrate that with personalization, our method has more capability of generalization; the results show that our model can deal with scenarios in which the source images are out of the domain of training set and even when the source images are in cartoon style from the Internet.}
	\label{fig:ab_personalization} 
\end{figure*}
We perform the ablation studies of with/without personalization to verify the effectiveness of personalization.  Besides, we also analyze the effect of hyper-parameters, including the number of source images $1\le s_n \le 8$ and that of steps $T$ for personalization. Since we only use the Youtube-Dancer-18 dataset in the testing phase, it is reasonable to evaluate the generalization of our methods of with/without personalization on this dataset. Here, we use self-imitation evaluation metrics, as mentioned above.

\begin{table}[h]
	\centering
	\caption{Comparison of our proposed AttLWB with and without personalization on the Youtube-Dancer-18 dataset. $\uparrow$ means the larger the better and $\downarrow$ means the smaller the better.}
	\begin{tabular}{c|c|c|c|c|c|c}
		\hline
		\multicolumn{1}{m{0.1cm}<{\centering}|}{ }        & \multicolumn{1}{m{0.8cm}<{\centering}|}{PSRN$\uparrow$}     & \multicolumn{1}{m{0.8cm}<{\centering}|}{SSIM$\uparrow$}  & \multicolumn{1}{m{0.8cm}<{\centering}|}{LPIPS$\downarrow$}    & \multicolumn{1}{m{1.0cm}<{\centering}|}{Body-CS$\uparrow$} & \multicolumn{1}{m{1.0cm}<{\centering}|}{Face-CS$\uparrow$} & \multicolumn{1}{m{0.8cm}<{\centering}}{FID$\downarrow$}     \\ \hline
		w/o   & 16.932 & 0.519 & 0.302 & 0.792     & 0.335  & 79.321 \\
		\textbf{with}  & \textbf{17.974} & \textbf{0.579}   & \textbf{0.263} & \textbf{0.834}  & \textbf{0.413}   & \textbf{59.832} \\ \hline
	\end{tabular}
	\label{table:personalization}
\end{table}

\textbf{With/Without Personalization.}  We conduct comparative experiments with and without personalization in our methods. Here, we fix the $s_n=2$ and $T=100$ in the phase of personalization. Table~\ref{table:personalization} shows that our method with personalization could achieve $1.0421$ higher in PSNR, $0.0599$ higher in SSIM, and $0.039$ lower in LPIPS than that without personalization on the Youtube-Dancer-18 dataset. Furthermore, we display some example results in Fig.~\ref{fig:ab_personalization}, where the left-column two roles are the source images from the Youtube-Dancer-18 dataset, and the right-column two roles are the cartoon images from the Internet. We find that with personalization, 1) our method could keep the color style of the background unchanged, as shown in the $1^{st}$ top left of Fig.~\ref{fig:ab_personalization}; 2) our method is capable of preserving the logo details in the source clothes, as depicted in the $2^{nd}$ bottom left of Fig.~\ref{fig:ab_personalization}; 3) our method is more powerful in the generalization, even when the source images are cartoon style, as illustrated in the right column of Fig.~\ref{fig:ab_personalization}. These demonstrate that personalization indeed plays a significant role in improving the generalization of our system.

\begin{figure}[h]
	\centering
	\includegraphics[width=\linewidth]{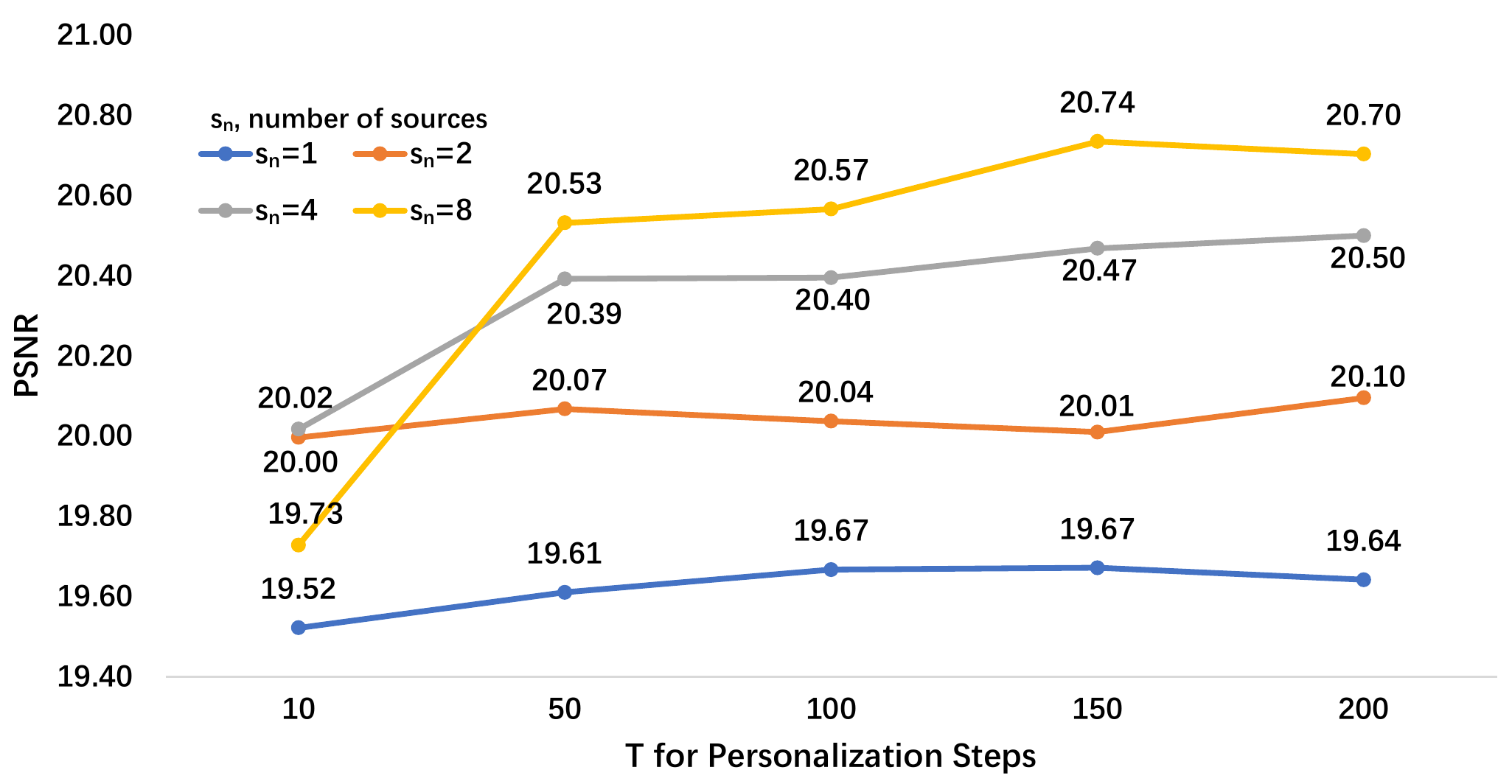}
	\caption{Comparison of different number of source images $s_n$ and number of steps $T$ for personalization. The performance grows with the increase of $s_n$, when $T$ is large enough. When $T$ is small with respective to a large $s_n$, in this case of $T=10$ and $s_n=8$, the performance would decrease.}
	\label{fig:ns_T} 
\end{figure}

\textbf{Number of Source Images $s_n$.}
In our system, we adopt a few source images $1\le s_n \le 8$ for personalization, and we will analyze the impacts of $s_n$ to the final results. Here, we fix the number of steps to $T=t \in \{10, 50, 100, 150, 200\}$ respectively for personalization and list the PSNR with different $s_n \in \{1, 2, 4, 8\}$ in Fig.~\ref{fig:ns_T}. It shows that the performance grows with an increase of $s_n$ when $T$ is large enough. The reason for the performance increase is due to the increase of the invisible textures. However, it is worth noticing that when $ T $ is small with respect to a large $s_n$, in the case of $T=10$ and $s_n=8$, the performance decreases. The reason might be that when $T$ is small, it is too hard for the network to fit those too many source images.

\textbf{Number of Steps $T$ for Personalization.} In the real application, we should take the number of steps $T$ into consideration because more steps will take more time. It is necessary to consider the trade-off between performance and overhead time for personalization. We set $s_n \in \{1, 2, 4, 8\}$ and list the performance with different $T$ for personalization in Fig.~\ref{fig:ns_T}. From Fig.~\ref{fig:ns_T}, we can see that the performance saturates at around 150 steps.

In summary, based on the above analysis, we recommend that in the stage of personalization, finetuning around 100 steps should be enough, and if the time for personalization is limited, it would be better to use fewer source images.

\subsection{Impact of Different Loss Functions}
In our methods, we apply a perceptual loss $L_p$, a face identity loss $L_f$, an attention regularization loss $L_a$, and an adversarial loss $L^G_{adv}$ (with global, body and head adversarial loss in details) to the full training loss functions. To validate the effectiveness of each term, we perform the ablation studies of the different loss functions. From Table~\ref{table:loss_function}, we can see that the model with the full loss would have the best performance. Besides, with the addition of $L_f$ and $L^G_{adv}$, the performance increases compared with that of the trial with only $L_p$.

\begin{table}[h]
	\centering
	\caption{Comparison between results with different loss functions on the Youtube-Dancer-18 dataset.$\uparrow$ means the larger the better and $\downarrow$ means the smaller the better.}
	\begin{tabular}{c|c|c|c|c|c}
		\hline
		\multicolumn{1}{m{0.8cm}<{\centering}|}{ } & 	\multicolumn{1}{m{0.8cm}<{\centering}|}{ PSNR$\uparrow$}           & \multicolumn{1}{m{0.8cm}<{\centering}|}{SSIM$\uparrow$}           & \multicolumn{1}{m{0.8cm}<{\centering}|}{LPIPS$\downarrow$}          & \multicolumn{1}{m{0.8cm}<{\centering}|}{Body-CS$\uparrow$}     & \multicolumn{1}{m{0.8cm}<{\centering}}{Face-CS$\uparrow$}        \\ \hline
		$L_p$                                        & 18.204          & 0.575          & 0.274          & 0.791          & 0.314          \\
		$L_p+L^G_{adv}$                              & 19.656          & 0.638          & 0.231          & 0.810          & 0.334          \\
		$L_p+L^G_{adv} + L_f$ & 19.542          & 0.629          & 0.247          & 0.809          & 0.351          \\
		\textbf{$L_{full}$}                          & \textbf{20.038} & \textbf{0.656} & \textbf{0.212} & \textbf{0.826} & \textbf{0.421} \\ \hline
	\end{tabular}
	\label{table:loss_function}
\end{table}

\subsection{Impact of Different Warping Strategies}
To verify the impact of our proposed Attentional Liquid Warping Block (AttLWB), we design some baselines with the ways mentioned above to propagate the source information, including input concatenation, texture warping, and feature warping. The body recovery, flow composition modules, the basic network architectures, and all loss functions are the same except for the propagating strategies among our method and other warping baselines. Here, we denote early concatenation, texture warping, and feature warping, as $ W_C $, $ W_T $, and $ W_F $, respectively. Also, we denote the $s_n$ source images as $\{I_{s_1}, ..., I_{s_n}\}$, their corresponding conditional inputs as $\{C_{s_1}, ..., C_{s_n}\}$ and their corresponding feature maps as $\{X^l_{s_1}, ..., X^l_{s_n}\}$ at the $l^{th}$ layer, respectively. The reference conditional inputs are $C_t$. The transformation flow of each source image to the reference is $T_{s_i\to t}$. We list the details of all warping baselines in followings:

\textbf{Input Concatenation $W_C$}. It directly concatenates all source images, their corresponding conditional inputs, as well as the reference conditional inputs, and then feeds them into the $G_{TSF}$ network, as shown in Fig.~\ref{fig:fusion} (a).

\textbf{Texture Warping $W_T$}. Based on each transformation flow $T_{s_i\to t}$, we warp each source image $s_i$ to the reference condition, average the pixels of overlap regions, and synthesize an initial image. Then, we feed it into the $G_{TSF}$ network and generate the final image, as shown in Fig.~\ref{fig:fusion} (b).

\textbf{Feature Warping $W_F$}. Instead of warping the source information in the image space, it propagates the source information in the feature space, based on the transformation flow. As mentioned above, we firstly obtain the warped feature $X^l_{s_i\to t}$ by using a bilinear sampler (BS) to warp each source feature $X^l_{s_i}$ concerning the corresponding transformation flow $T_{s_i\to t}$. According to the ways to aggregate the global feature $X^l_t$ from multiple warped source features $\{X^l_{s_1\to t}, ..., X^l_{s_n\to t}\}$, we can specifically subdivide them into the followings:

\begin{enumerate}
	\item \textbf{Attention $W^{Att}_F$} (ours) is shown in Algorithm~\ref{alg:attlwb}.
	\item \textbf{Add-Aggregation $W^{A}_F$} (ours). It is the first version of our proposed Liquid Warping Block(LWB)~\cite{Liu_2019_ICCV}, as shown in the Fig.~\ref{fig:lwb} (a) and Equation~(\ref{equ:addlwb}).
	
	\item \textbf{Mean-Aggregation $W^{M}_F$}. Directly adding the warped features will enlarge the magnitude of the features in the overlap area and thereby results in artifacts. A naive way is to average all the warped features, shown as follows.
	
	\begin{equation}
	\begin{aligned}
	\widehat{X}_t^{l} &= \frac{1}{s_n}\sum_{i=1}^{s_n}X^{l}_{s_i \to t} + X_t^{l}.
	\end{aligned}
	\label{equ:mean_lwb}
	\end{equation}
	
	\item \textbf{Add-Soft-Gate $W^{A\odot}_F$}. The warped feature might introduce the misalignment problem, and to address it, Dong~\etal~\cite{softgate18} utilizes a gated convolution to control the transformation degree. We firstly add all the warped features, then utilize a gated convolution, as shown in Equation~(\ref{equ:soft_gate_add_lwb}). Here, $g$ is a function with two-convolution layers followed by a Sigmoid activation and $g(X_t^{l})\in [0, 1]$. $\odot$ represents the element-wise multiplication.
	
	\begin{equation}
	\begin{aligned}
	\widehat{X}_t^{l} &= g(X_t^{l})\odot \sum_{i=1}^{s_n}X^{l}_{s_i \to t} + X_t^{l}.
	\end{aligned}
	\label{equ:soft_gate_add_lwb}
	\end{equation}
	
	\item \textbf{Mean-Soft-Gate $W^{M\odot}_F$}. It firstly averages all the warped features and following steps are the same with $W^{A\odot}_F$. The formulation is shown as follows:
	\begin{equation}
	\begin{aligned}
	\widehat{X}_t^{l} &= g(X_t^{l})\odot \frac{1}{s_n}\sum_{i=1}^{s_n}X^{l}_{s_i \to t} + X_t^{l}.
	\end{aligned}
	\label{equ:soft_gate_mean_lwb}
	\end{equation}
\end{enumerate}

\begin{figure}[h]
	\centering
	\includegraphics[width=0.8\linewidth]{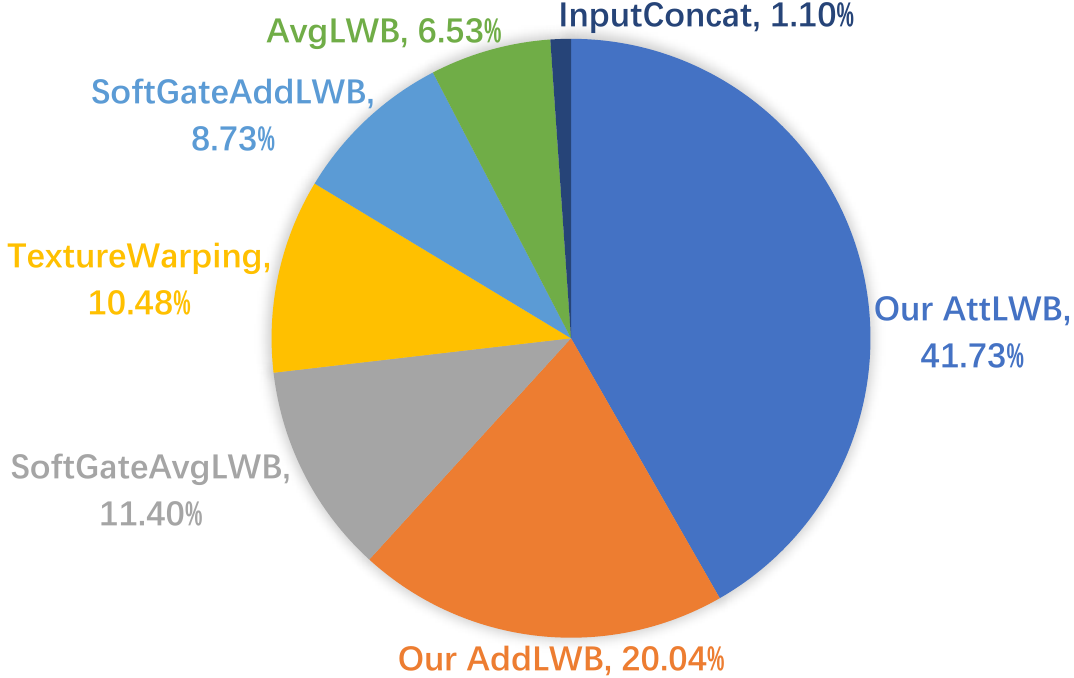}
	\caption{Results of the user study (\%). The user preference of the videos with best quality regarding to the quality of face, the quality of clothes texture and background.}
	\label{fig:user_study} 
\end{figure}

We conduct a user study, with 64 volunteers, to assess the quality of the generated videos and compare the performance of the warping strategies mentioned above. Participants are shown 17 groups of videos with 7 videos generated by 7 warping strategies respectively in random order in each group. Besides, the shared source image and reference video of each group is also shown to the participants for reference. Participants are asked to choose the best video considering the quality of the face, clothes texture, figure pose, and background. Finally, 64 responses are collected, and the results are shown in Fig.~\ref{fig:user_study}. As we can see that our proposed AttLWB and AddLWB have the best appraise, scoring $41.73\%$ and $20.04\%$, respectively, far higher than others.

\subsection{Running Time}
\begin{figure}[h]
	\centering
	\includegraphics[width=0.9\linewidth]{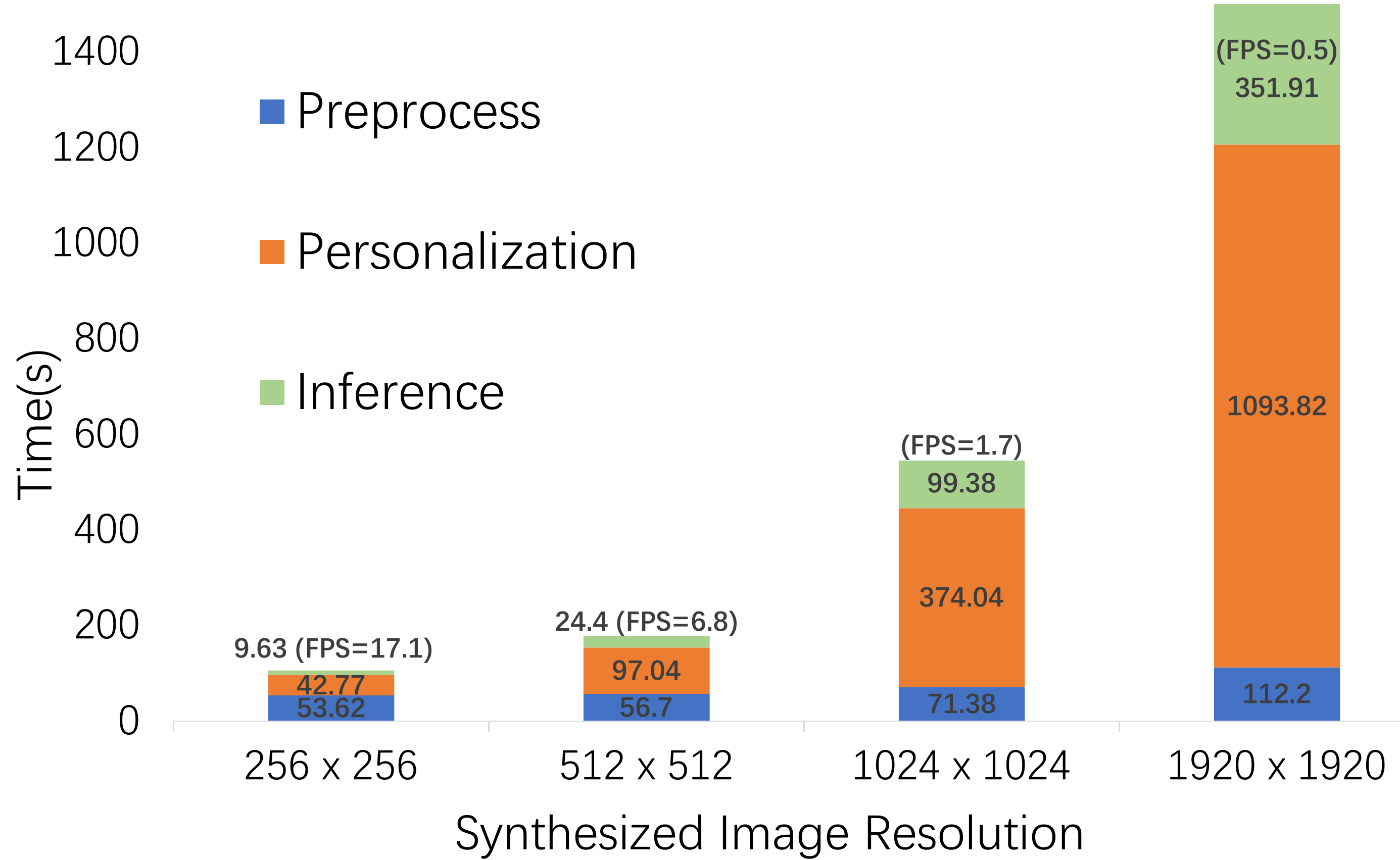}
	\caption{Running time when producing images with different resolutions. The I/O consumption has been taken into count. The larger resolution, the more consuming time is, particularly in the stages of personalization and inference. }
	\label{fig:running_time} 
\end{figure}

Our method could produce the results with different image resolutions, ranging from $256\times256$, $512\times512$, $1024\times1024$ to $1920\times1920$. Here, we benchmark the running time of our system in different image resolutions. Since a high resolution needs more memory allocation of the GPUs device, we perform all the tests on a Tesla V100S-PCIe-32G GPU with the Intel Xeon(R) E5-2620 2.10GHz CPUs. The image resolution of the source images is $4032\times3024$, and that of the reference video with 165 frames is $1920 \times 1080$. In Fig.\ref{fig:running_time}, we separately report the running time of preprocessing, personalization and inference, when synthesizing different resolutions, respectively. From Fig.\ref{fig:running_time}, we can see that the higher resolution consumes more running time, especially in the personalization and inference. 

\subsection{Failure Cases and Limitations}
\begin{figure}[h]
	\centering
	\includegraphics[width=0.9\linewidth]{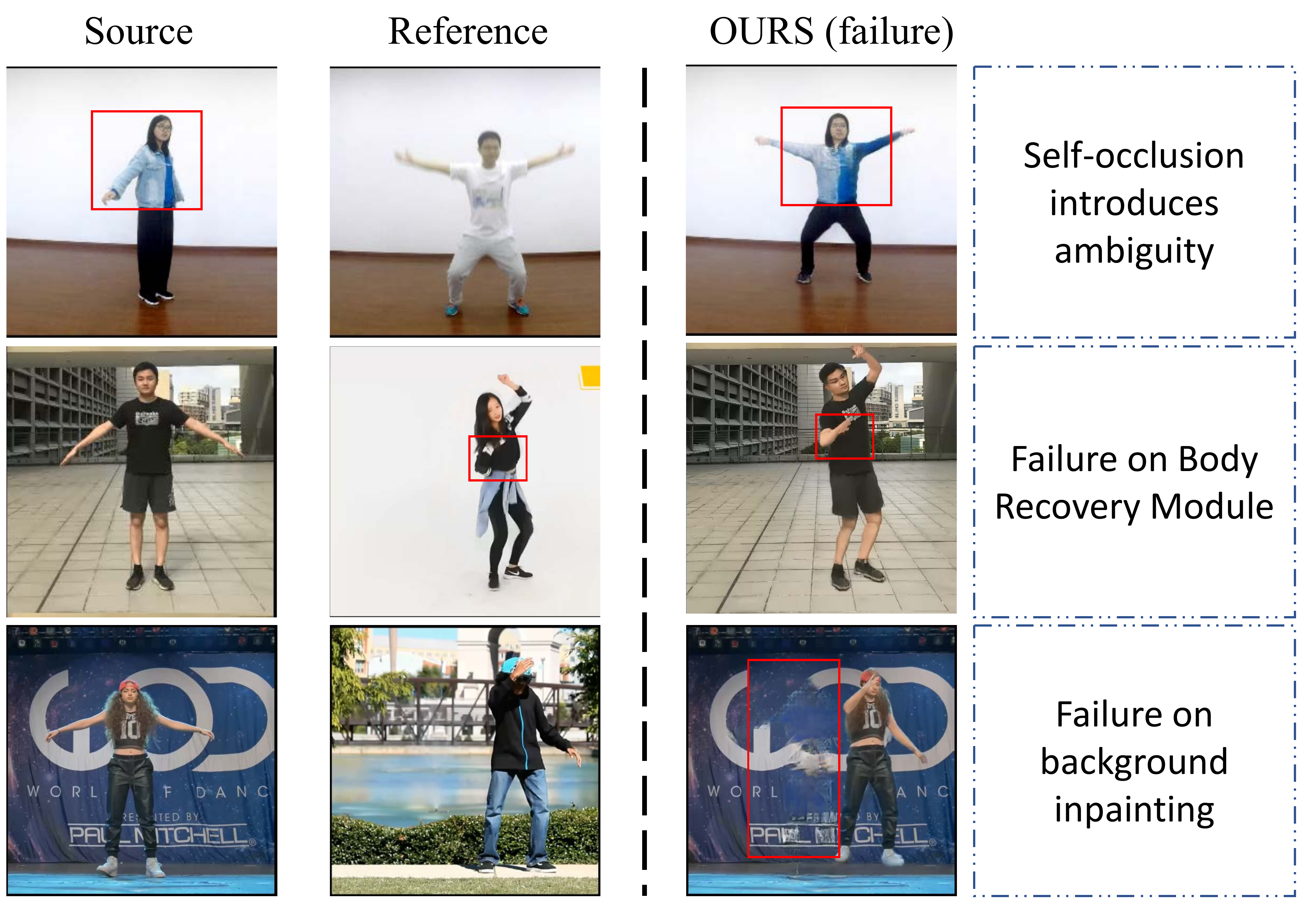}
	\caption{The failure cases of our system. It mainly contains three types of failure cases. One occurs when the source images introduce a large self-occlusion area, as shown in the top row. The second row is when the body recovery module fails. The third row shows the artifacts when the background inpainting network fails.}
	\label{fig:failures} 
\end{figure}

There are three main types of failure cases of our methods.
The first one, as shown in the $1^{st}$ row of Fig.~\ref{fig:failures}, is that source image contains a large area of self-occlusion, which introduces an ambiguity in textures and thereby results in a bad synthesized image. The second occurs when the Body Recovery Module fails and could not accurately estimate the pose parameters, as illustrated in the $2^{nd}$ row of Fig.~\ref{fig:failures}. The rest is when the background inpaintor $G_{BG}$ fails, as shown in the $3^{rd}$ row of Fig.~\ref{fig:failures}. 

In addition, there are still some limitations of our system, 1) it cannot imitate the motions of hands and facial expressions from the reference images, since the 3D body parametric SMPL~\cite{SMPL:2015} used in our system does not contain the articulated hands and expressive face; 2) also, it cannot animate the large-motion body with too loose clothing like the skirt or evening dress; 3) it is affected by the different lighting environments among sources.

Therefore, for a better result, the input source images need to follow these guidelines:

\begin{itemize}
	\item They share the same static background without too complex scene structures. If possible, we recommend using the actual background.
	
	\item The person in the source images holds an A-pose for introducing the most visible textures.
	
	\item It is recommended to capture the source images in an environment without too much contrast in lighting conditions and lock auto-exposure and auto-focus of the camera.
\end{itemize}

\section{Conclusion}
We propose a unified framework to handle human motion imitation, appearance transfer, and novel view synthesis. It employs a body recovery module to estimate the 3D body mesh, which is more powerful than the 2D poses. In order to preserve the source information, we further design a novel warping strategy, Attentional Liquid Warping Block (AttLWB), which propagates the source information in both image and feature spaces and supports a more flexible warping from multiple sources. Besides, with a fast personalization, our method could be generalized well when the input images are out of the domain of training set and synthesize higher resolution ($512 \times 512$ and $1024 \times 1024$) results.  Extensive experiments show that our framework outperforms others and produce decent results.

\section*{ACKNOWLEDGMENT}
We thank Dr. Weixin Luo for the meaningful discussion in the whole procedure, and we appreciate all the help of building the first version of the iPER dataset from Min Jie.

\ifCLASSOPTIONcaptionsoff
  \newpage
\fi

\begin{IEEEbiography}[{\includegraphics[width=1in,height=1.25in,clip,keepaspectratio]{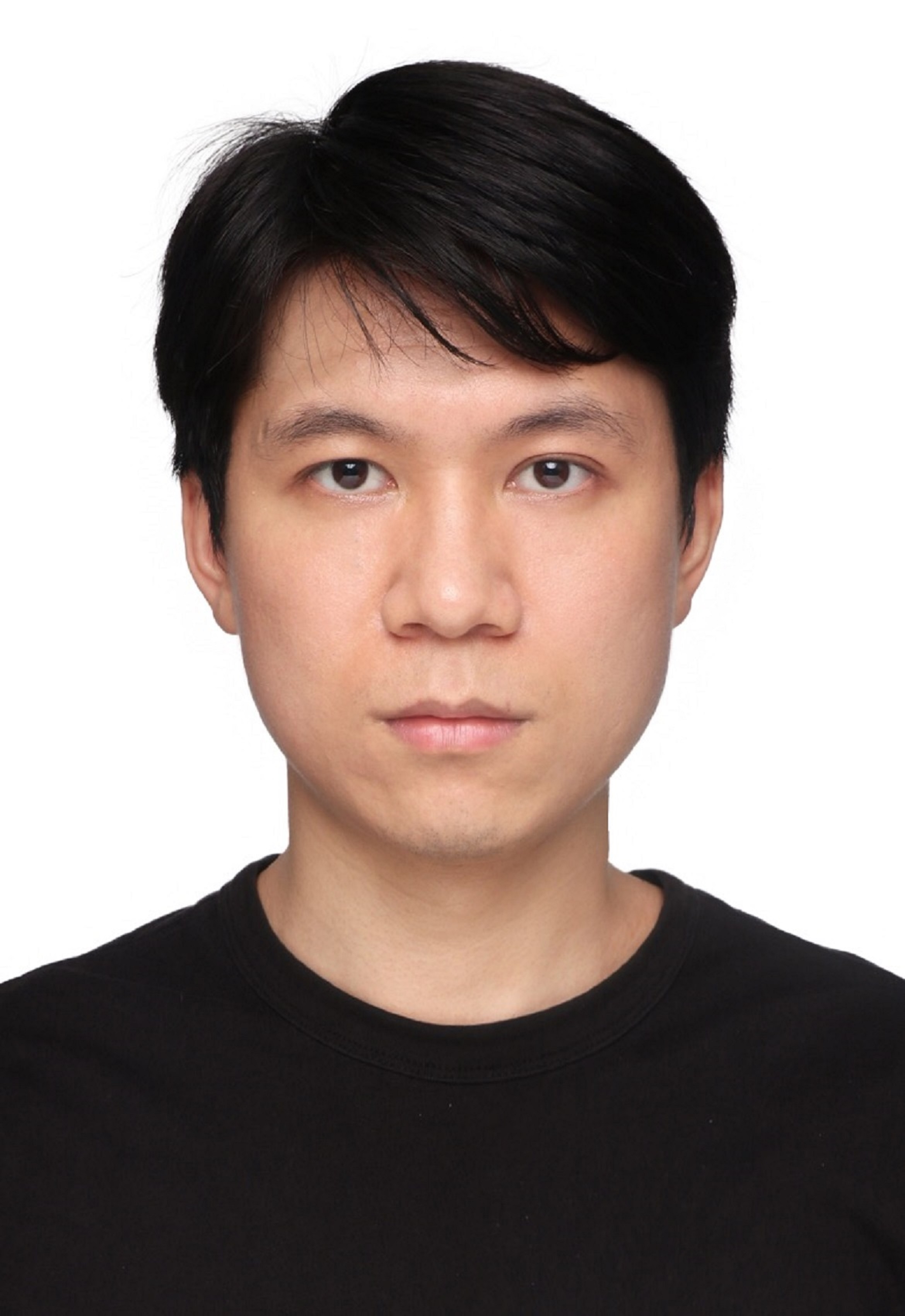}}]{Wen Liu} received the bachelor degree from Northwestern Polytechnical University, Xian, China, in 2016. He is currently pursuing a Ph.D. degree at ShanghaiTech University. His research interests focus on human 3D body reconstruction, image synthesis, motion transfer, novel view synthesis, neural rendering and video anomaly detection.
\end{IEEEbiography}

\begin{IEEEbiography}[{\includegraphics[width=1in,height=1.25in,clip,keepaspectratio]{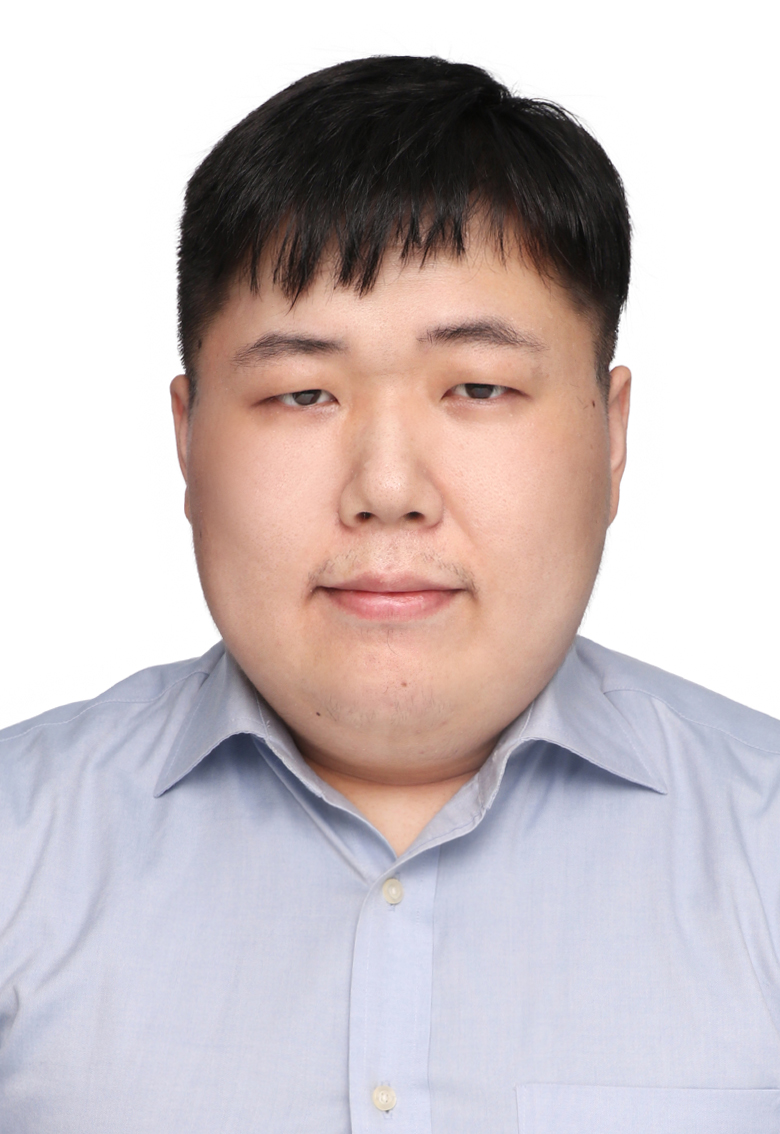}}]{Zhixin Piao} received the bachelor degree from Southeast University, Nanjing, China, in 2017. He is currently pursuing a master degree at ShanghaiTech University. His research topic is human 3D reconstruction and motion transfer.
\end{IEEEbiography}

\begin{IEEEbiography}[{\includegraphics[width=1in,height=1.25in,clip,keepaspectratio]{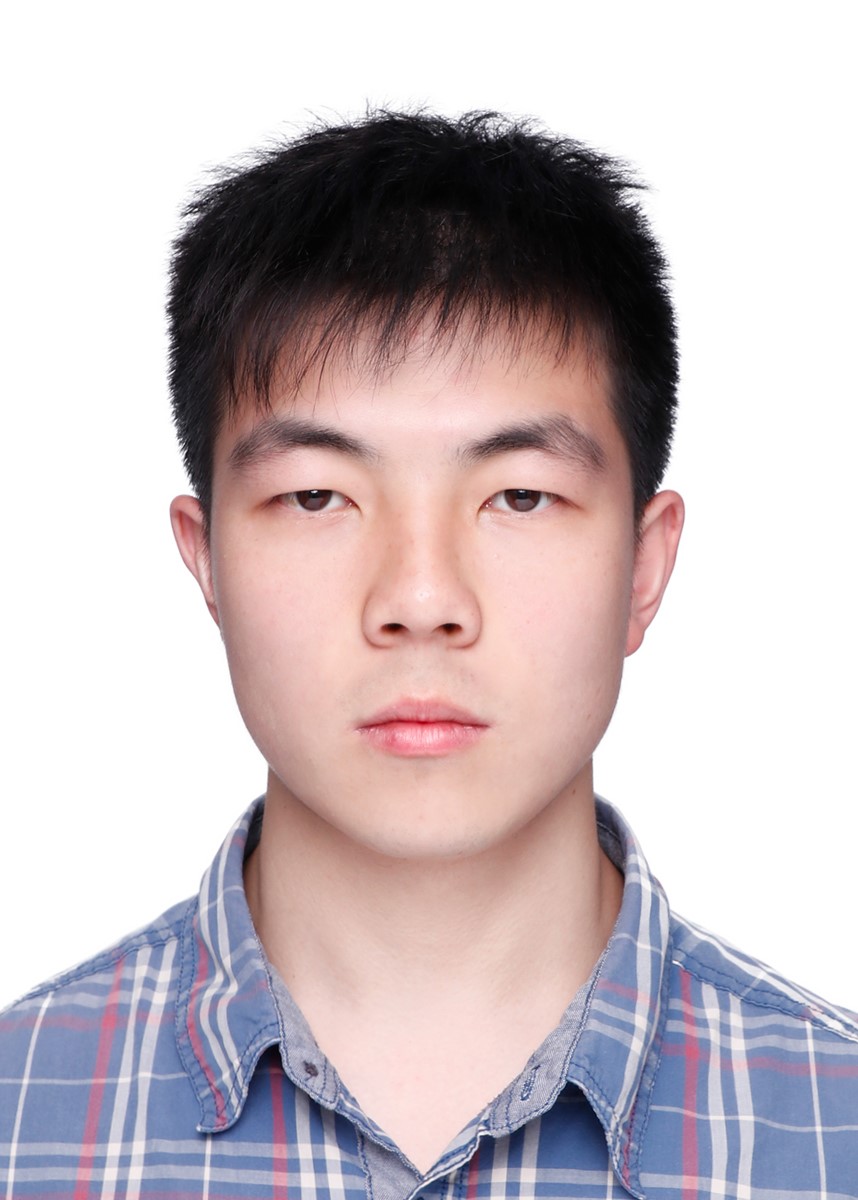}}]{Zhi Tu}
received the bachelor degree from ShanghaiTech University, Shanghai, China, in 2020. His research topic is human motion transfer and medical image analysis.
\end{IEEEbiography}

\begin{IEEEbiography}[{\includegraphics[width=1in,height=1.25in,clip,keepaspectratio]{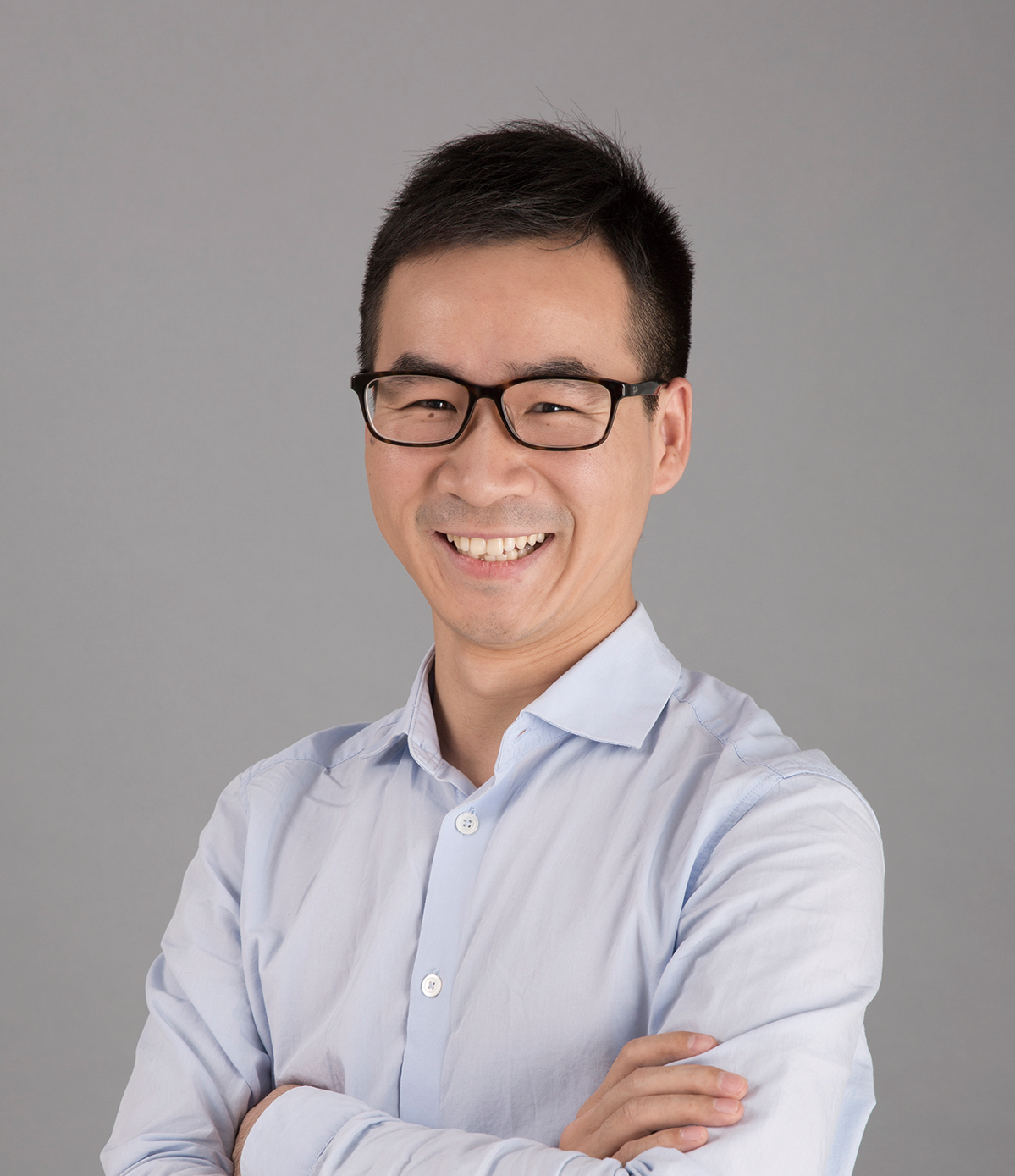}}]{Wenhan Luo} received the Ph.D. degree from Imperial College London, UK, 2016, M.E. degree from Institute of Automation, Chinese Academy of Sciences, China, 2012 and B.E. degree from Huazhong University of Science and Technology, China, 2009. His research interests include several topics in computer vision and machine learning, such as motion analysis (especially object tracking), image/video quality restoration, object detection and recognition, reinforcement learning.
\end{IEEEbiography}

\begin{IEEEbiography}[{\includegraphics[width=1in,height=1.25in,clip,keepaspectratio]{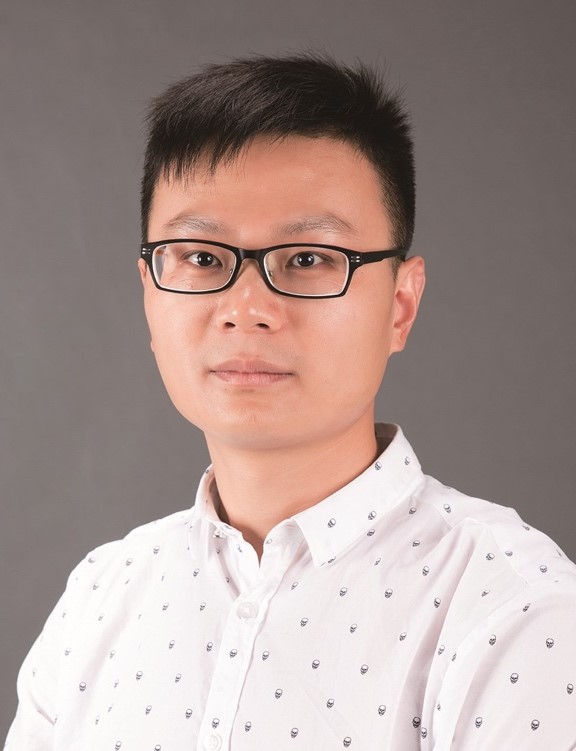}}]{Lin Ma} received the B.E. and M.E. degrees in computer science from the Harbin Institute of Technology, Harbin, China, in 2006 and 2008, respectively, and the Ph.D. degree from the Department of Electronic Engineering, The Chinese University of Hong Kong, in 2013. He was a Researcher with the Huawei Noah’s Ark Laboratory, Hong Kong, from 2013 to 2016. He is currently a Principal Researcher with the Tencent AI Laboratory, Shenzhen, China. His current research interests lie in the areas of computer vision, multimodal deep learning, specifically for image and language, image/video understanding, and quality assessment. Dr. Ma received the Best Paper Award from the Pacific-Rim Conference on Multimedia in 2008. He was a recipient of the Microsoft Research Asia Fellowship in 2011. He was a finalist in HKIS Young Scientist Award in engineering science in 2012.
	
\end{IEEEbiography}

\begin{IEEEbiography}[{\includegraphics[width=1in,height=1.25in,clip,keepaspectratio]{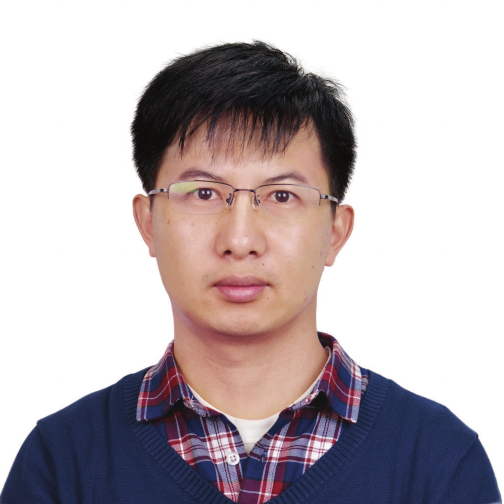}}]{Shenghua Gao} is an assistant professor, PI in ShanghaiTech University, China. He received the B.E. degree from the University of Science and Technology of China in 2008 (outstanding graduates), and received the Ph.D. degree from the Nanyang Technological University in 2012. From Jun 2012 to Jul 2014, he worked as a postdoctoral fellow in Advanced Digital Sciences Center, Singapore. His research interests include computer vision and machine learning.
\end{IEEEbiography}

\end{document}